\pdfoutput=1
\documentclass[11pt]{article}

\usepackage{cmap}          
\usepackage[T1]{fontenc}
\usepackage[utf8]{inputenc}
\usepackage[margin=1in]{geometry}
\usepackage{booktabs}
\usepackage{amsmath,amssymb}
\usepackage{graphicx}
\usepackage{xcolor}
\usepackage[hidelinks]{hyperref}
\hypersetup{pdftitle={Quantization Thresholds Replicate, Failure Modes Do Not: A Three-Model Study of Agentic Tool Use in Polish from 8-bit to 2-bit},pdfauthor={Jakub Prejzner}}
\makeatletter\g@addto@macro{\UrlBreaks}{\do\-}\makeatother   
\usepackage{multirow}
\usepackage{array}
\usepackage{caption}
\usepackage{placeins}
\usepackage{needspace}
\usepackage{pgfplots}
\usepackage{authblk}
\usepackage[protrusion=true,expansion=false]{microtype}
\pgfplotsset{compat=1.18}

\definecolor{c7b}{HTML}{1b6ca8}
\definecolor{c11b}{HTML}{9b2226}
\definecolor{cpll}{HTML}{2a7d4f}

\title{\textbf{Quantization Thresholds Replicate,\\ Failure Modes Do Not:\\
A Three-Model Study of Agentic Tool Use in Polish\\ from 8-bit to 2-bit}}

\author[1]{Jakub Prejzner\thanks{Corresponding author: rd4drop@gmail.com.}}
\affil[1]{Independent Researcher}

\date{September 25, 2026}

\begin{document}
\maketitle

\begin{abstract}
\noindent
We ask how GGUF quantization affects agentic tool use in Polish and whether the effects generalize across models. We introduce \textbf{PolAgentBench}, a deterministic benchmark with Polish prompts and English tool schemas: a 67-task main suite (15 adversarial probes, 52 hard-tier tasks) and a 46-task arithmetic isolation ladder. Three models span two axes of variation: Bielik-11B-v3.0 and its pruned, distilled child Bielik-Minitron-7B-v3.0 isolate model compression, and Llama-PLLuM-8B adds a change of pretraining family. Each is measured at six precisions, Q8\_0 to Q2\_K. Only the collapse threshold replicates. \textbf{(1)} All three models fall off a cliff between 3-bit and 2-bit (11B $0.716 \to 0.045$, 7B $0.463 \to 0.149$, PLLuM $0.224 \to 0.015$; paired McNemar $p<0.001$ in each), across a fourfold capability spread and both axes. \textbf{(2)} Failure modes do not replicate: at 2-bit the 7B fails long (median 9.1k tokens, 4 steps) while the 11B mostly answers at the first step with a confabulated final answer (37 of 64 failures); PLLuM fails on content across precisions (71.8--92.0\% of steps parse). \textbf{(3)} On the arithmetic ladder the unscaffolded rung is a floor, left standing by a rerun that states the no-tool rule; four explicit calls lift the 8-bit 11B from 1/10 to 9/10 and the 7B from 0/10 to 7/10 after format-only failures with the gold value are forgiven, an exploratory effect with eight distinct baseline inputs that does not survive multiplicity correction, while the order-trap arm separates the models at 8-bit (11B 6/6, 7B 0/6). \textbf{(4)} The Polish-versus-English gap is associated with degradation or with task family. We document four artifacts that shaped our conclusions (rounding-hostile gold values, strict answer typing, a no-tool rule the prompt never stated, priority-ordered failure labels), report affected results in strict and corrected form, and release the benchmark, trajectories and commit-stamped artifacts.
\end{abstract}

\section{Introduction}

Small open-weight language models are almost never deployed at full precision. A model that nominally requires 14 to 22\,GB in half precision is in practice shipped as a 4-bit or 3-bit GGUF and run on commodity hardware~\cite{llamacpp,quantsurvey}. Meanwhile the dominant use of such models has shifted from answering to \emph{acting}: calling tools, reading their outputs, and chaining further calls on the result. How quantization interacts with that second mode, and whether any answer transfers across models, is poorly characterized, and nearly all existing evidence concerns English.

\paragraph{Why small local models are the right subject.} A reasonable objection is that frontier hosted models now solve agentic tasks well, so the behaviour of a quantized 7B is of narrow interest. We disagree for four reasons, each describing a setting where a hosted model is not an option rather than merely a more expensive one. Confidentiality constraints in legal, medical, and public-sector work often forbid sending documents to an external API at all, so the practical question is not which model is best but whether any locally runnable model suffices. Air-gapped and intermittently connected deployments have the same property. At high volume, the per-call cost difference between a local quantized model and a hosted frontier model determines whether a product exists. And for languages outside the English-centric mainstream, locally deployable national models are a matter of technological sovereignty as much as of cost. In all four settings the operative question is exactly ours: how far can this model be compressed before it stops working as an agent, and how will it fail when it does.

We study that question for Polish, under a deployment condition that is both realistic and demanding. The user request arrives in Polish while the tool schema, meaning function names, parameter names, and enum values, is English. The model must sustain a continuous semantic translation while emitting valid structured output. Polish adds morphological pressure: case declension, where the locative \emph{Łodzi} must be normalized to the nominative \emph{Łódź} before it is a valid argument, and numerals spelled out in oblique cases, where \emph{osiemdziesięciu} must be parsed to 80.

Our original hypothesis (H1) was that quantization degrades agentic ability asymmetrically earlier than general quality, so that a model might look acceptable on a standard benchmark at 4-bit while already failing as an agent. We did not measure the general-quality side of that comparison, so H1 remains untested here; what the agentic curves alone show is that none of them erodes early and monotonically, and pursuing the question surfaced something more interesting. Because we ran an identical suite on three models that vary along two separable axes, compression within one family and a change of family at comparable scale, we can ask which observations are properties of \emph{quantization} and which are properties of \emph{a particular model}. That separation organizes the paper, and it cuts sharply: of four candidate findings, exactly one replicates.

\begin{itemize}
  \item \textbf{RQ1.} Where does agentic ability collapse under quantization, and is the location model-dependent?
  \item \textbf{RQ2.} How do models fail as precision drops, and is the failure mode itself stable across models?
  \item \textbf{RQ3.} Is arithmetic over tool outputs a capability floor, and does tool scaffolding help or hurt?
  \item \textbf{RQ4.} Does the Polish and English interface split impose a measurable cost, and is that cost stable?
\end{itemize}

\noindent\textbf{Contributions.} We present PolAgentBench, a deterministic and commit-stamped benchmark of Polish-interface agentic tool use, comprising a 67-task main suite and a 46-task arithmetic isolation ladder with two four-call arms, one carrying an order trap and one free of it, together with a length by language design that breaks the confound between chain length and interface language. We report a three-model, six-precision measurement showing that the 3-bit to 2-bit collapse threshold is shared while failure modes, the reasons each model sits on the arithmetic floor, survival of the order trap, and language gaps are not. We give a mechanistic account of one model's dominant high-precision failure as a repairable protocol defect, supported by trajectory-level evidence and raw model output. Finally we document four methodological hazards that materially changed our own conclusions: a rounding artifact in synthetic gold values, a strict answer-typing rule that penalized correct computations, and a no-tool rule that the oracle enforced but the prompt never stated, numbered as Artifacts 1 to 3 in Section~\ref{sec:mines}, together with a fourth, the priority-ordered failure taxonomy of Section~\ref{sec:taxonomy}, whose labels are less determinate than they appear; affected ladder results are reported in strict and corrected form, the third hazard is measured by a rerun on the extended ladder, and its residual exposure in the main suite is stated in Section~\ref{sec:mines}.

\section{Related Work}

\paragraph{GGUF quantization and its evaluation.} The GGUF $k$-quant family (Q2\_K through Q8\_0) used here is the de facto format for local inference~\cite{llamacpp}, built on block-wise uniform quantization with superblock refinements at low bit-widths; surveys of resource-efficient inference cover the broader design space~\cite{quantsurvey}. Closest to our measurement setting is a unified evaluation of llama.cpp quantization on Llama-3.1-8B-Instruct~\cite{kquanteval}, which asks how much quality is lost between $k$-quants and whether 3-bit variants remain usable downstream. We extend that question along three axes it does not cover: a non-English interface, agentic rather than static tasks, and multiple models. Beyond llama.cpp, low-bit studies of the Llama-3 family document severe degradation at 2 to 3 bits on static benchmarks~\cite{huang2024llama3}, and a systematic study of quantized reasoning models finds that 4-bit weights are close to lossless while lower bit-widths hurt, increasingly so as tasks get harder~\cite{liu2025reasoning}. Closest to our multilingual question is Marchisio et al.~\cite{marchisio2024}, who show that quantization harms non-English performance more than English, non-Latin scripts most, and that automatic metrics understate the damage that human raters see; they measure static tasks across languages, whereas we measure agentic trajectories in one language under a mixed-language interface.

\paragraph{Tool-use and agentic benchmarks.} The Berkeley Function Calling Leaderboard is the standard reference for function-calling evaluation, spanning simple, parallel, and stateful multi-turn settings~\cite{bfcl}. $\tau$-bench evaluates end-to-end completion in multi-step domains~\cite{taubench}, AgentBench spans multiple interactive environments~\cite{agentbench}, and ToolLLM scales tool-use training and evaluation to large real-world API collections~\cite{toolllm}. These are predominantly English and weight single-call correctness heavily. Methodologically our hard tier is closest to FuncBenchGen~\cite{funcbenchgen}, which formalizes multi-step function calling as traversal over a dependency graph that the model must infer rather than receive. We share the synthetic, contamination-free, controllable-difficulty design and differ in that our controlled variable is quantization and our interface is non-English. TinyLLM~\cite{tinyllm} is closest in motivation, covering small models on agentic tasks for edge deployment, but traces no quantization curve and does not address non-English interfaces.

\paragraph{Polish models.} The Bielik family targets Polish across several scales and generations~\cite{bielik7b,bielik11bv2,bielikv3}. The two checkpoints evaluated here have their own reports: Bielik 11B v3 extends Mistral 7B v0.2 to 11B parameters by depth up-scaling~\cite{bielik11bv3}, and Bielik-Minitron-7B is derived from it by structured pruning and logit distillation in the NVIDIA Minitron style~\cite{bielikminitron,minitron}; the released checkpoints are documented in their model cards~\cite{bielik11bv3card,bielikminitroncard}. PLLuM is a separate Polish national effort producing Polish-adapted models on external lineages; we evaluate its Llama-3.1-based 8B instruct release~\cite{pllum}. This work follows the present author's earlier study of low-bit quantization of Bielik~\cite{priorwork}, which motivated the low-bit focus but measured general quality on a single model rather than agentic trajectories across models.

\section{PolAgentBench}
\label{sec:bench}

\subsection{Environment}
\begin{sloppypar}\noindent All tasks run in one deterministic weather environment exposing five tools: \texttt{get\_\allowbreak{}weather(city)}, \texttt{get\_\allowbreak{}forecast(city,\allowbreak{} days)}, \texttt{send\_\allowbreak{}weather\_\allowbreak{}alert(city,\allowbreak{} severity,\allowbreak{} message)}, \texttt{convert\_\allowbreak{}temperature(value,\allowbreak{} from\_\allowbreak{}unit,\allowbreak{} to\_\allowbreak{}unit)}, and \texttt{find\_\allowbreak{}nearest\_\allowbreak{}city(reference\_\allowbreak{}city,\allowbreak{} max\_\allowbreak{}distance\_\allowbreak{}km)}. The environment has no randomness, no clock, and no I/O. Every tool is a pure function of a static 52-city database (32 Polish, 20 international) plus a hand-built neighbour graph over nine Polish source cities. \texttt{get\_forecast} returns a deterministic ramp of $0.5\,^{\circ}$C per day that passes through the city's base temperature (on day 3 of a four-day forecast). Full environment determinism makes every tool output exactly reproducible, so that any variation between runs of the same task originates in decoding rather than in the environment, and localizes failures to the model. Appendix~\ref{app:example} gives a complete worked task with its gold trajectory.\end{sloppypar}

\subsection{Task tiers}
The \textbf{easy tier} (15 tasks) is a set of single-call and few-call adversarial probes targeting specific failure modes: Polish leaking into English enum values, inflection and diacritic handling, hallucinated tool results, unauthorized side effects, and over-calling. Per-task outcomes for the Bieliks are in Appendix~\ref{app:easy}.

The \textbf{hard tier} (52 tasks) consists of 40 chains of 2 to 6 steps in which each step's argument derives from a prior step's output, plus the 12-task pilot ladder of Section~\ref{sec:ladder}, whose lowest rungs need zero or one call. A recurring device is the \textbf{order trap}: \texttt{find\_nearest\_city} returns cities in graph order rather than distance order, so a chain instructing the model to use the \emph{first} returned city separates models that read tool output literally from those that substitute a plausible-sounding answer.

\subsection{The arithmetic isolation ladder}
\label{sec:ladder}
The ladder separates inability to compute from inability to compute while orchestrating. Within an instance, all rungs share an \textbf{identical gold value}; what varies is how much tool scaffolding leads to it. L0 supplies the numbers directly in the prompt and its oracle rejects any tool call, L1 requires one call, L2 two. The L0 prompt as run did not state that prohibition; Section~\ref{sec:mines} records this as Artifact 3 and Section~\ref{sec:f3} reports a rerun, L0e, in which the sentence is present.

The ladder exists in two stages. A 12-task \emph{pilot} (three instances per rung, the top rung being a four-tool chain with the order trap) is embedded in the main suite and produced the pattern that motivated everything else. Because three instances per rung admit a minimum attainable $p$ of only $0.25$, we built a 46-task \emph{extended} ladder as a separate suite: L0, L1, L2 at ten instance slots each, and two four-call arms. L0 and its rerun L0e each contain only eight distinct inputs: instances \texttt{a} and \texttt{e}, and \texttt{c} and \texttt{f}, share the prompt, tools and step budget (gold values $46.4$ and $45.5$). L1, L2 and L3N have ten distinct inputs each, and L3T six; Appendix~\ref{app:ladder} reports paired contrasts with \texttt{e} and \texttt{f} removed from both rungs. \textbf{L3T} ($n=6$) keeps the order trap and reuses the pilot's three trap routes, which we disclose; the graph's nine source cities admit no more genuinely distinct trap routes. \textbf{L3N} ($n=10$) removes the graph entirely: \texttt{get\_weather(A)}, then \texttt{get\_forecast(A,4)}, then two separate \texttt{convert\_temperature} calls on days 2 and 4, with the model averaging the two Fahrenheit values in its final answer. Because Celsius-to-Fahrenheit conversion is affine, converting then averaging equals averaging then converting, so the gold value is identical across every rung of an instance; we verified this equality on the live environment for all ten cities. The L3N specification requires both conversions as separate calls. That choice runs \emph{against} our own hypothesis: L3N adds calls to the same gold value relative to the lower rungs, so by design it should weaken rather than create the gradient we test for. This is a construction argument, not a tested property. Counting all ten slots gives a smallest attainable exact McNemar $p$ of $0.002$ with ten discordants and $0.0039$ with nine. After duplicate exclusion the minimum is $2/2^8=0.0078$; Section~\ref{sec:f3} states what this implies for the analysis-stage family of contrasts.

\subsection{Matched twins}
\label{sec:twins}
Ten hard-tier chains in the main suite exist as matched twin pairs. Both members share prompt, tool set, and required call sequence, and differ only in whether the oracle additionally requires the correct computed number in the final answer. This isolates value correctness from orchestration correctness on trajectories that share a prompt and, at $T=0$, usually but not always coincide (Section~\ref{sec:mines}). Three of these pairs (\texttt{v3\_chain\_001}, \texttt{004}, \texttt{005}) additionally share prompt, tools, budget and value oracle with the pilot ladder's top-rung tasks \texttt{v3\_arith\_L3\_a} to \texttt{c}, so the main suite holds 54 distinct model inputs in 67 slots; the $n=47$ sensitivity analyses contain no repeated input.

\subsection{Length and language, deconfounded}
\label{sec:axes}
\begin{sloppypar}\noindent Hard-tier chains vary in length (2 to 6 calls) and in interface language (\texttt{PL\_EN} for a Polish prompt with English schema, \texttt{EN\_EN} for both English). An earlier version of this suite confounded the two. Because the neighbour graph contains only Polish cities, \texttt{EN\_EN} chains could not use \texttt{find\_nearest\_city} and were therefore structurally shorter, making any language comparison partly a length comparison. We removed the confound by constructing long \texttt{EN\_EN} chains from graph-free tools and short \texttt{PL\_EN} chains, filling all four length by language cells.\end{sloppypar}

\subsection{Oracle, determinism, and three artifacts}
\label{sec:mines}
Evaluation uses a deterministic \texttt{evaluate()} with per-key oracles: ordered-subsequence checks for tool sequences and substring checks for computed answers, accepting both Polish comma and English decimal-point conventions but applying no rounding tolerance. Any step the parser rejects also fails the task, even when the model recovers and every task-level check passes (5 of the 11B's 19 failures at Q3\_K\_M carry only such step-level tags).

\paragraph{Artifact 1: the rounding hazard.} Exact matching interacts dangerously with synthetic gold values. In an earlier revision, gold averages ended in inconvenient decimals such as $7.76$, so a model that rounded a legitimate intermediate result to one decimal place produced a wrong final string and was scored as a failure despite computing correctly. In the first full run this contaminated 14 of 96 arithmetic failure slots. We rebuilt every arithmetic gold value using a construction in which the sampled forecast offsets cancel exactly, so that averaging day 2 and day 4 of a four-day forecast returns the city's base temperature exactly (every base temperature in the database is a multiple of $0.5\,^{\circ}$C, so its Fahrenheit value has at most one decimal) and each gold value is invariant to intermediate rounding. Each task was then validated three ways: the exact trajectory passes, a trajectory that rounds the intermediate also passes, and a wrong-value trajectory still fails. After correction the artifact fires exactly once across the whole grid, at the 7B's Q5\_K\_M point (Appendix~\ref{app:taxonomy}), and is zero at every other precision in all three models. The consequence was not cosmetic: in the contaminated run the collapse threshold appeared to sit at the 4-bit to 3-bit transition ($p=0.0215$, measured on the archived pre-correction run, which is excluded from the release); on corrected data that contrast is not significant on any tier and the real cliff sits one step lower. The pre-correction result would not have replicated.

\begin{sloppypar}\paragraph{Artifact 2: answer typing, noticed once the runs were under way.} The final-answer schema requires a string. On the extended ladder we found failures in which the model makes the required calls in order, arrives at the correct number, and then returns it as a bare float (\verb|{"answer": 46.4}|, typical of the 11B) or as an object (\verb|{"answer":| \verb|{"average_temperature_fahrenheit": 46.4}}|, typical of the 7B). The schema validator rejects the payload, no final answer is registered, and the task fails with \texttt{final\_answer\_missing} and \texttt{schema\_violation} tags despite the correct value. Unlike the rounding hazard, this artifact surfaced only once the ladder runs were under way (the released run log flags it after the first 11B ladder cells and before the 7B ladder runs), not before the decisive runs, so the oracle is left unmodified and we adopt a \emph{sensitivity protocol} instead: every affected result is reported at two rates, \textbf{strict} (the oracle as run) and \textbf{corrected} (forgiving only failures that carry format tags alone, with no tool-sequence or content tag, and whose mistyped answer contains the gold value; for the 7B this also forgives a rejected extra call before the answer, an attempt to convert its Fahrenheit mean once more as if it were Celsius, so the 7B's corrected rate is an upper bound, Appendix~\ref{app:ladder}). The correction is conservative by construction: the 31 L1 and L2 failures of the 11B at Q8\_0 and Q4\_K\_M carrying the same tags but substantively wrong values (for example $59$ where the gold is $46.4$) are not forgiven. A pre-specified second sensitivity split, for models that shortcut the two required conversions into one, turned out to be empty: no such trajectory exists in the data. The typing artifact, not the anticipated shortcut, was the real hazard. Section~\ref{sec:f3} reports both rates; Appendix~\ref{app:ladder} lists every corrected cell.\end{sloppypar}

\paragraph{Artifact 3: a rule stated to the oracle but not to the model, found by audit.} Every task file carries a \texttt{constraints} field of design notes, and the L0 rungs of both ladders, the ten extended-ladder tasks and the three pilot tasks inside the main suite, additionally carry the oracle key \texttt{no\_tool\_calls} (as does the over-calling probe \texttt{adv\_010}, by design). Neither reaches the model: the runner builds its messages from the system prompt and the task prompt alone. The L0 prompt therefore asked for an average without saying that tools were forbidden, while the oracle failed any trajectory that used one; the 11B, offered five tools, used them in every L0 trajectory at Q8\_0, Q4\_K\_M and Q3\_K\_M and was scored as if it had disobeyed an instruction it never saw. A post-release audit found this in 2026-09. As with the typing artifact we leave the runs untouched and add a measurement instead: L0 was rerun with the prohibition stated in the prompt (L0e) on the four cells the scaffolding observation rests on, together with a control that reproduced the original raw model outputs, token counts and verdicts (Section~\ref{sec:f3}, Appendix~\ref{app:ladder}). The three pilot L0 tasks inside the main suite were not rerun: on them the 11B is failed on \texttt{unexpected\_tool\_call} alone, its final answer carrying the exact gold value, in four main-suite slots (one at Q8\_0, one at Q6\_K, two at Q3\_K\_M, at both repair settings), so its main-suite counts at those precisions are strict counts that a stated rule could have raised to 55, 57 and 50 of 67; no 7B or PLLuM slot on these three tasks fails on that tag alone. No conclusion of Section~\ref{sec:f1} changes: the cliff contrasts could only widen, and the Q4\_K\_M dip would deepen (nominal $p=0.013$ against Q3, Holm-adjusted $0.075$; $0.064$ and $0.38$ without twins) without surviving correction. The \texttt{constraints} field stays in the task files as documentation and is labelled as such in the repository.

\paragraph{Decoding determinism, and what was actually verified.} Before the runs, greedy decoding was checked on one single-call smoke task across two model-load cycles and found identical in raw output and parsed action at every step; on that basis the mainline uses a single seed (42) at temperature 0 throughout, and Section~\ref{sec:variance} probes what happens under sampling. That check is narrower than the guarantee a reader might infer from it, and the released data show the difference. The matched twins of Section~\ref{sec:twins} send the model an identical prompt, tool set, and step budget, so at $T=0$ the two members should produce the same trajectory; in fact they coincide in step count and token count in 46 of 60 twin pairs for the 11B, 32 of 60 for the 7B, and 31 of 60 for PLLuM (six precisions by ten pairs each, \texttt{repair\,=\,off}). Judged on the raw model output rather than on step and token count, 45, 28 and 27 pairs are identical (100 of 180); where they diverge, the outputs differ from the first step in 43 of those 80 pairs and later in the other 37, one member dropping the call envelope or taking a different route to the same tools. This is consistent with floating-point nondeterminism in CUDA inference, and we do not attribute it further. It is not the same as run-to-run variation: the one rerun we have, the L0 control of Section~\ref{sec:f3}, reproduced all 80 of its trajectories exactly under a different driver and CUDA version, so the divergence may depend on where in a run a prompt is decoded rather than on chance. Two consequences follow. Greedy decoding within a run is reproducible for most but not all tasks, so every cell of the grid is a single draw from a decoder whose output depends on more than the prompt; the sampling probe of Section~\ref{sec:variance} shows how wide the spread becomes when randomness is added deliberately. And twin pairs are strongly correlated observations of one prompt rather than two independent ones; the main-suite tests of Section~\ref{sec:f1} and Appendix~\ref{app:mcnemar} nevertheless count all 67 task slots, both members of each pair included, because they were built before the divergence was measured, so Appendices~\ref{app:mcnemar} and~\ref{app:bootstrap} repeat the decisive contrasts and intervals with both members of each pair removed ($n=47$); Section~\ref{sec:twins-results} reads the twin contrast in the light of the divergence.

Every hard-tier task was validated offline before any GPU run by constructing an ideal trajectory from real tool outputs and confirming that it passes the oracle while a give-up trajectory fails.

\subsection{Failure taxonomy, and what its labels do and do not mean}
\label{sec:taxonomy}

Section~\ref{sec:f2} rests on a classification of failures into three modes. Because the classifier's design shapes the result, we state it here.

The taxonomy applies to failing trajectories on the 25 arithmetic-graded hard-tier tasks at \texttt{repair\,=\,off}. It is fully automatic and deterministic, with no manual adjudication. It reads only the oracle tags of the stored trajectory, the final answer text, and the task's gold values. Rules are checked in order and the first match wins:

\begin{sloppypar}
\begin{enumerate}
  \item If the tag set contains \texttt{wrong\_final\_answer}, enter the arithmetic branch and ignore all other tags. Within it, extract every number from the answer; if any lies within $0.06$ of a gold value, label \textsc{rounding artifact}, otherwise \textsc{genuine arithmetic}.
  \item Otherwise, if the tag set intersects \{\texttt{invalid\_\allowbreak{}json}, \texttt{no\_\allowbreak{}json\_\allowbreak{}found}, \texttt{schema\_\allowbreak{}violation}, \texttt{unknown\_\allowbreak{}action}, \texttt{final\_\allowbreak{}answer\_\allowbreak{}shape\_\allowbreak{}violation}\}, label \textsc{protocol shape}.
  \item Otherwise, if the tag set intersects \{\texttt{final\_\allowbreak{}answer\_\allowbreak{}missing}, \texttt{wrong\_\allowbreak{}tool\_\allowbreak{}order}, \texttt{unexpected\_\allowbreak{}tool\_\allowbreak{}call}\}, label \textsc{tool fixation}.
  \item Otherwise, \textsc{other}.
\end{enumerate}
\end{sloppypar}

\paragraph{The labels are less determinate than they look.} The taxonomy is single-label with a hard priority order, and many failures carry tags from more than one category, so the priority rule decides a large share of assignments. Table~\ref{tab:purity} reports how often a failure's tags fall within a single category, which we treat as a label-confidence indicator. Two consequences follow and are applied consistently in Section~\ref{sec:f2}. The 3-bit labels are the most trustworthy for the Bieliks (13 of 20 and 8 of 10 single-category). The 7B's high-precision plateau and the 2-bit labels for both Bieliks are the least trustworthy. At 7B 6-bit not one of the 21 failures is single-category and at 8-bit only 1 of 21 is; both cells break the same way, 15 of 21 and 18 of 21 respectively carrying protocol and fixation tags together, so the \textsc{protocol shape} count records which rule fired first rather than establishing that fixation was absent. PLLuM's labels are no better determined: only 39 of its 140 failures across the curve trigger a single rule branch, the dominant co-occurrence, a wrong final answer together with fixation tags, covers 90 of 140, and reversing the priority would relabel 101 of the 140, that is 72\% of all PLLuM failures (Appendix~\ref{app:taxonomy}); PLLuM's failure profile in Section~\ref{sec:f2} therefore rests on priority-independent tag counts, not on the single-label winner.

\begin{table}[h]
\centering\small
\caption{Label determinacy on the 25 arithmetic-graded tasks, \texttt{repair\,=\,off}: failures whose tags all fall within one taxonomy class. Full co-occurrence counts in Appendix~\ref{app:taxonomy}.}
\label{tab:purity}
\begin{tabular}{lccccccc}
\toprule
& & Q8\_0 & Q6\_K & Q5\_K\_M & Q4\_K\_M & Q3\_K\_M & Q2\_K \\
\midrule
\multirow{2}{*}{7B}  & failures      & 21 & 21 & 21 & 21 & 20 & 25 \\
                     & single category & 1 & \textbf{0} & 4 & 5 & \textbf{13} & 7 \\
\midrule
\multirow{2}{*}{11B} & failures      & 11 & 10 & 11 & 17 & 10 & 25 \\
                     & single category & 3 & 3 & 4 & 2 & \textbf{8} & \textbf{3} \\
\midrule
\multirow{2}{*}{PLLuM} & failures    & 23 & 23 & 24 & 23 & 22 & 25 \\
                     & single category & 6 & 9 & 6 & 4 & 9 & 5 \\
\bottomrule
\end{tabular}
\end{table}

\paragraph{That co-occurrence is a signature, not noise.} The heavy overlap of protocol and fixation tags at 7B 8-bit is exactly what the defect of Section~\ref{sec:envelope} produces. When the model writes a tool name into the action field, the parser rejects the step, which raises a protocol tag, and the tool therefore never executes, so the required call never appears in order, which raises a fixation tag. One defect, two tags. The priority rule assigns it to protocol, which matches the mechanism, but the assignment is made by fiat rather than by evidence internal to the classifier. A multi-label count that refuses to force a winner finds tags from more than one class in most 2-bit ladder failures (35 of 46 for the 11B, 30 of 46 for the 7B), so no 2-bit claim in this paper rests on the taxonomy; 2-bit claims rest on trajectory cost, which is classifier-independent.

\paragraph{Coverage and verification.} Four oracle tags observed at \texttt{repair\,=\,off} belong to no category and would fall to \textsc{other}; on the 25-task subset \textsc{other} is empty for all model by precision combinations. We reimplemented the rules from this description and reproduced the published counts exactly.

\section{Experimental Setup}
\label{sec:setup}

We evaluate three models at six GGUF precisions each (Q8\_0, Q6\_K, Q5\_K\_M, Q4\_K\_M, Q3\_K\_M, Q2\_K):

\begin{itemize}
  \item \textbf{Bielik-11B-v3.0-Instruct}~\cite{bielik11bv3,bielik11bv3card}, Mistral 7B v0.2 depth-upscaled to 11B parameters. All six quantizations obtained pre-built from a single community publisher (\texttt{DevQuasar}), so that one static quantization recipe covers the whole curve.
  \item \textbf{Bielik-Minitron-7B-v3.0-Instruct}~\cite{bielikminitron,bielikminitroncard}. This is \emph{not} an independent model: it is produced from Bielik-11B-v3-Base by structured pruning (11.04B to 7.35B parameters, NVIDIA Minitron-style~\cite{minitron}) followed by logit distillation from the parent, and it shares the parent's architecture family, training corpus, and tokenizer~\cite{bielikminitron}. Q8\_0, Q6\_K, Q5\_K\_M, and Q4\_K\_M were obtained directly from the publisher's official GGUF release; because that release stops at Q4\_K\_M and no public half-precision GGUF exists, Q3\_K\_M and Q2\_K were requantized from Q8\_0 via \texttt{llama-quantize -{}-allow-requantize}.
  \item \textbf{Llama-PLLuM-8B-instruct}~\cite{pllum}, a Polish-adapted model from the PLLuM national consortium built on the Llama-3.1-8B lineage, run with the \texttt{llama-3} chat template; GGUF conversions from a community publisher (\texttt{mradermacher}).
\end{itemize}

\paragraph{Two axes instead of one confound.} An earlier draft of this work treated the Bielik pair as differing in scale and architecture. The parent-child relation above narrows that: the pair shares its lineage, corpus and tokenizer and differs by structured pruning of depth and width (11.04B to 7.35B parameters), logit distillation and the child's own post-training, so what it isolates is \emph{model compression as a package}, with the architecture family held fixed rather than architecture and scale held fixed. PLLuM supplies the second axis, a change of pretraining family and corpus at comparable scale, while remaining a Polish-focused model, so the family axis is not confounded with the interface language. A result that replicates across both axes, compression within a family and a change of family, is hard to attribute to anything but quantization itself. A result that differs between the Bieliks may be an effect of compression rather than of parameter count, and we phrase all such results accordingly.

\paragraph{PLLuM provenance, disclosed.} The evaluation protocol included a pre-specified sanity gate: at 8-bit the model must reach 0.5 on the easy tier before the full curve is run. PLLuM scored $0.400$ ($6/15$) and the gate failed; per protocol the phase was skipped, and the curve was subsequently executed in remaining compute budget and is reported here with that provenance. Before using the data we checked for harness faults: PLLuM emits well-formed protocol steps at a rate comparable to the 11B (83.9\% of steps parse at 8-bit, against the 11B's 89.3\%), its transcripts show no chat-template artifacts, and a ceiling analysis that forgives every failure touched by any format-related tag still caps it at $0.478$, far below the 11B's actual $0.806$. Its failures are dominated by content: summing the tag occurrences recorded in \texttt{summary.json}, with \texttt{final\_answer\_missing} counted as a format tag, the 8-bit run has 74 content tags (every tag outside the format set, \texttt{timeout} included) against 29 format tags, the inverse of the 8-bit 11B's 15 against 43. These checks support reading PLLuM as a weak tool-use model on this suite, while leaving prompt compatibility as a limitation. We therefore use it for within-model contrasts, above all the location of the collapse threshold, and for its failure-mode profile, and we do not use it for between-model capability comparisons.

\paragraph{Infrastructure.} Inference used \texttt{llama-cpp-python} 0.3.19 (prebuilt CUDA-12 wheel) on a single NVIDIA RTX 4090 with full GPU offload, temperature 0, seed 42, at most 512 completion tokens per step, a step budget set per task, ChatML formatting for the Bieliks and \texttt{llama-3} for PLLuM. Main-suite budgets range from 4 to 10 steps and are 8 on 29 of the 67 tasks (4 steps on 4 tasks, 5 on 9, 6 on 14, 10 on 11); every task of the extended ladder is given 8. The context window was 8192 tokens for all models. Every run writes a \texttt{summary.json} stamped with the repository commit hash. Independent oracle re-evaluation of stored trajectories reproduces every reported pass count exactly for the Bielik main-suite runs (24 of 24). Main-suite Bielik results carry commits \texttt{83db813} (7B) and \texttt{a023c3b} (11B); the extended ladder, the PLLuM curve, the 7B grid completion, and the sampling probe carry commit \texttt{e584b38}. These are development identifiers; the public repository holds byte-identical trees as \texttt{0b46390}, \texttt{6d71818} and \texttt{56607b6} (\texttt{docs/commit\_map.tsv}).

\paragraph{Statistics.} Paired McNemar exact tests (two-sided) for contrasts between precisions on the same tasks and, on the ladder, between rungs of the same instances, the instance being the pairing unit because all rungs of an instance share one gold value. Task-resampling bootstrap confidence intervals with 10{,}000 resamples; paired differences use a paired bootstrap over tasks. Twin pairs enter the main-suite tests as separate task slots; Appendix~\ref{app:mcnemar} gives the sensitivity with both members removed. Unless noted, analyses are for \texttt{repair\,=\,off}.

\section{Finding 1: The collapse threshold replicates in all three models}
\label{sec:f1}

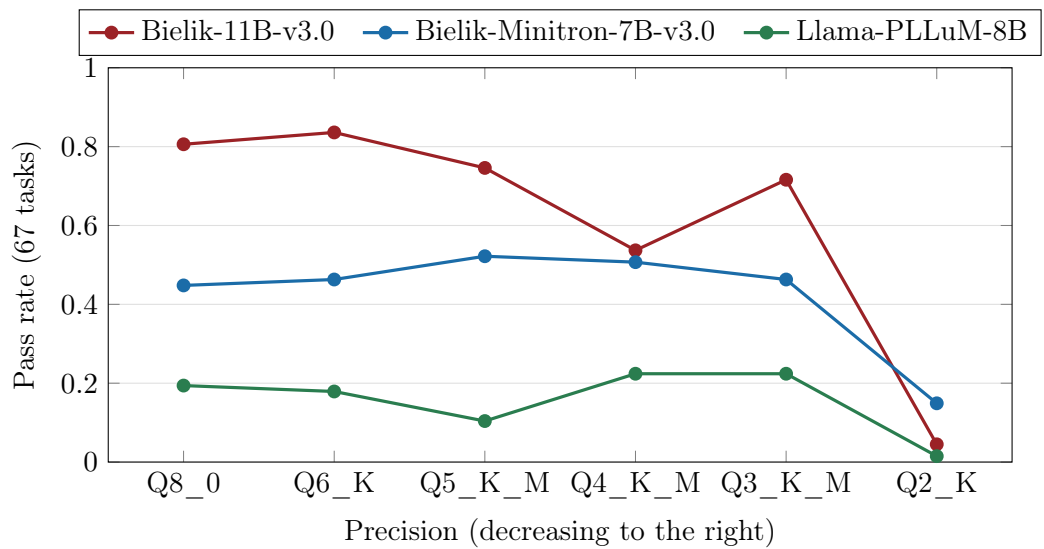
\begin{figure}[h]
\centering
\begin{tikzpicture}
\begin{axis}[
  width=0.82\textwidth, height=6.8cm,
  xlabel={Precision (decreasing to the right)},
  ylabel={Pass rate (67 tasks)},
  xtick={1,2,3,4,5,6},
  xticklabels={Q8\_0,Q6\_K,Q5\_K\_M,Q4\_K\_M,Q3\_K\_M,Q2\_K},
  ymin=0, ymax=1, ymajorgrids, grid style={gray!25},
  legend style={at={(0.5,1.03)},anchor=south,legend columns=-1,/tikz/every even column/.append style={column sep=8pt}}, legend cell align=left,
  every axis plot/.append style={very thick, mark=*},
]
\addplot[c11b] coordinates {(1,0.806)(2,0.836)(3,0.746)(4,0.537)(5,0.716)(6,0.045)};
\addlegendentry{Bielik-11B-v3.0}
\addplot[c7b] coordinates {(1,0.448)(2,0.463)(3,0.522)(4,0.507)(5,0.463)(6,0.149)};
\addlegendentry{Bielik-Minitron-7B-v3.0}
\addplot[cpll] coordinates {(1,0.194)(2,0.179)(3,0.104)(4,0.224)(5,0.224)(6,0.015)};
\addlegendentry{Llama-PLLuM-8B}
\end{axis}
\end{tikzpicture}
\caption{Quantization curves, \texttt{repair\,=\,off}. Three very different curves, one shared terminus: every model ends in the same cliff between 3-bit and 2-bit.}
\label{fig:curve}
\end{figure}

Table~\ref{tab:curves} gives the three quantization curves and Figure~\ref{fig:curve} plots them on a shared precision axis.

\begin{table}[h]
\centering
\caption{All-task pass rate (67 tasks) by model and precision, \texttt{repair\,=\,off} and \texttt{on}. All three models fall off a cliff between Q3\_K\_M and Q2\_K. Tier-resolved counts in Appendix~\ref{app:split}.}
\label{tab:curves}
\begin{tabular}{lcccccc}
\toprule
& Q8\_0 & Q6\_K & Q5\_K\_M & Q4\_K\_M & Q3\_K\_M & Q2\_K \\
\midrule
7B    & 0.448\,/\,0.567 & 0.463\,/\,0.567 & 0.522\,/\,0.597 & 0.507\,/\,0.597 & 0.463\,/\,0.463 & 0.149\,/\,0.179 \\
11B   & 0.806\,/\,0.806 & 0.836\,/\,0.851 & 0.746\,/\,0.761 & 0.537\,/\,0.597 & 0.716\,/\,0.716 & 0.045\,/\,0.075 \\
PLLuM & 0.194\,/\,0.224 & 0.179\,/\,0.209 & 0.104\,/\,0.134 & 0.224\,/\,0.254 & 0.224\,/\,0.269 & 0.015\,/\,0.015 \\
\bottomrule
\end{tabular}
\end{table}

All three models survive substantial compression and then fail abruptly in the same place. On the hard tier the 11B drops from $36/52$ to $1/52$ ($p<0.0001$) and the 7B from $18/52$ to $2/52$ ($p<0.0001$). PLLuM drops from $0.224$ to $0.015$ on all tasks, with 14 discordant tasks against 0 ($p=0.00012$) and a paired bootstrap difference of $+0.209$, 95\% interval $[+0.119, +0.313]$. Full contrast tables are in Appendix~\ref{app:mcnemar}, bootstrap intervals in Appendix~\ref{app:bootstrap}.

What makes this claim strong is what it survives. The three models span a fourfold range in absolute capability at 8-bit ($0.194$ to $0.806$). The Bielik pair differs by structured pruning and distillation; PLLuM differs by pretraining family, corpus, and tokenizer; the low-bit quantizations come from three different provenances. As the next three sections show, essentially every other behaviour we measured differs across these models. The threshold location is the one thing that does not. We therefore read it as a property of $k$-quantization at 2 to 3 bits rather than of any model, with the qualification that all three subjects are Polish-focused models in the 7B to 11B range.

Three secondary observations follow. First, no curve is monotone, and no two are non-monotone in the same way. The 11B's nominal peak is Q6\_K, and its Q4\_K\_M point is a nominal local dip below both neighbours ($p=0.0043$ against Q5, $p=0.029$ against Q3, uncorrected and with twins counted); with both members of each twin pair removed the contrast against Q3 is no longer significant ($p=0.14$) and, after this exclusion followed by Holm correction over the fourteen 11B contrasts, neither neighbour is (adjusted $p=0.27$ and $0.86$; with twins counted, $0.030$ and $0.17$) (Appendix~\ref{app:mcnemar}), so we describe the dip rather than claim it; it recovers at 3-bit, and we have no mechanistic account of it. The 7B's curve is a statistical plateau from Q8\_0 through Q3\_K\_M: its nominal maximum sits at Q5\_K\_M ($0.522$), but no pairwise contrast above the cliff is significant (Q5 against Q4, $p=1.0$ with thirteen discordant tasks, an informative null; Q6 against Q8, $p=1.0$ with a single discordant task, a comparison that could not have detected anything), so we do not claim a peak, only a flat usable band ending in the cliff. PLLuM moves inside a low band ($0.104$ to $0.224$) with a nominal dip at Q5\_K\_M that we report descriptively. Second, the ordering of the Bieliks reverses at the extreme: the 11B exceeds the 7B at every precision except Q2\_K, where it collapses further ($0.045$ against $0.149$). Within this family, capability buys headroom across the usable range and costs graceful degradation at the extreme. Third, for a model that is weak at the task to begin with, precision is visibly not the binding constraint: PLLuM's curve is nearly flat until the universal cliff removes what little there was.

\section{Finding 2: Failure modes do not replicate; three models, three signatures}
\label{sec:f2}

Pass rates say where models fail; they do not say how. Table~\ref{tab:modes} gives the failure classification of Section~\ref{sec:taxonomy} for all three models, to be read together with the determinacy figures of Table~\ref{tab:purity}.

\begin{table}[h]
\centering
\caption{Failure modes on the 25 arithmetic-graded tasks, \texttt{repair\,=\,off}: total failures, then \textsc{protocol shape}, \textsc{genuine arithmetic}, \textsc{tool fixation}. The rounding artifact fires exactly once across the whole grid, at the 7B's Q5\_K\_M point, and is zero in every other cell; that one case is why the three mode counts in that cell sum to 20 rather than 21 (Appendix~\ref{app:taxonomy}). Labels on the 7B's 8-bit to 5-bit plateau, at 2-bit everywhere, and throughout the PLLuM row are priority-dominated and indicative only (Section~\ref{sec:taxonomy}).}
\label{tab:modes}
\setlength{\tabcolsep}{4pt}%
\resizebox{\linewidth}{!}{%
\begin{tabular}{lcccccc}
\toprule
& Q8\_0 & Q6\_K & Q5\_K\_M & Q4\_K\_M & Q3\_K\_M & Q2\_K \\
\midrule
7B    & 21 / 16 / 4 / 1 & 21 / 12 / 9 / 0 & 21 / 12 / 6 / 2 & 21 / 7 / \textbf{13} / 1 & 20 / 5 / 4 / \textbf{11} & 25 / 6 / 15 / 4 \\
11B   & 11 / 5 / 4 / 2 & 10 / 3 / 5 / 2 & 11 / 4 / 6 / 1 & 17 / 9 / 7 / 1 & 10 / \textbf{0} / 7 / 3 & 25 / 5 / 19 / 1 \\
PLLuM & 23 / 0 / \textbf{23} / 0 & 23 / 0 / 22 / 1 & 24 / 4 / 20 / 0 & 23 / 1 / 22 / 0 & 22 / 1 / 20 / 1 & 25 / 0 / \textbf{23} / 2 \\
\bottomrule
\end{tabular}%
}
\end{table}

Taken at face value the 7B exhibits an orderly progression: protocol violations dominate across the whole 8-bit to 5-bit plateau, genuine miscalculation at 4-bit, fixation at 3-bit. The 3-bit fixation result is solid (13 of 20 failures single-category, corroborated by trajectory cost below); the protocol result across the 8-bit to 5-bit plateau is priority-dependent, and least determined of all at 6-bit, though Section~\ref{sec:envelope} identifies a concrete mechanism behind it. \textbf{The 11B does not reproduce the progression}: protocol and arithmetic failures interleave across its healthy precisions, its 3-bit protocol failures fall to zero in the cell where labels are most determinate, and its fixation counts never rise the way the 7B's do. \textbf{PLLuM shows a third profile}, and it is where the single-label taxonomy must be read most carefully. Under the as-run priority nearly every PLLuM failure lands in genuine arithmetic (130 of 140 across the whole curve; protocol shape 6, fixation 4), at every precision including 2-bit. That label is priority-dominated: 90 of the 140 carry a wrong final answer together with fixation tags, chiefly \texttt{wrong\_tool\_order}, so a reversed priority would call most of them fixation instead (Appendix~\ref{app:taxonomy}). What survives either reading is the tag-level picture of Section~\ref{sec:setup}: format tags are rare (29 against 74 content tag occurrences at 8-bit, the inverse of the 11B), parse fluency is of the same order as the 11B's (71.8\% to 92.0\% of steps parse across the six precisions, against 61.3\% to 93.6\% for the 11B), and the ceiling under maximal format forgiveness is $0.478$ at 8-bit and at most $0.507$ at any precision. Read either way, PLLuM fails on the task itself, wrong values and wrong call structure, while speaking the protocol, and its 2-bit collapse adds no new signature; it simply removes the few successes there were. That is the third profile: not a different way of breaking under quantization, but the same content failure at every precision, terminated by the shared cliff.

\subsection{Two opposite collapses in the Bielik family, visible in trajectory cost}
The clearest divergence is at the shared threshold, and it is measured without reference to the taxonomy. Table~\ref{tab:tokens} contrasts the token and step cost of failing against passing trajectories for the Bieliks; token counts are comparable within this pair because both models share one tokenizer, and cross-family comparisons with PLLuM would require steps rather than tokens.

\begin{table}[h]
\centering
\caption{Median tokens and steps per trajectory, split by outcome, Bielik pair, \texttt{repair\,=\,off}. Tokens are prompt plus completion tokens summed over all completion calls of a trajectory, so the conversation context is counted again at every step. Step budgets are per task, 4 to 10, and are 8 on 29 of the 67 tasks. Full distributions in Appendix~\ref{app:tokens}.}
\label{tab:tokens}
\begin{tabular}{llcccc}
\toprule
& & \multicolumn{2}{c}{median tokens} & \multicolumn{2}{c}{median steps} \\
\cmidrule(lr){3-4}\cmidrule(lr){5-6}
model & precision & pass & fail & pass & fail \\
\midrule
\multirow{6}{*}{7B}
 & Q8\_0   & 3504 & 12298 & 3.0 & 8.0 \\
 & Q6\_K   & 3515 & 12384 & 3.0 & 8.0 \\
 & Q5\_K\_M & 3577 & 14065.5 & 3.0 & 8.0 \\
 & Q4\_K\_M & 3501 & 10793 & 3.0 & 6.0 \\
 & Q3\_K\_M & 3509 & 26833 & 3.0 & 8.0 \\
 & Q2\_K   & 3230.5 & \textbf{9139} & 2.0 & \textbf{4.0} \\
\midrule
\multirow{6}{*}{11B}
 & Q8\_0   & 5914 & 8415 & 4.5 & 6.0 \\
 & Q6\_K   & 6378.5 & 7510 & 4.5 & 5.0 \\
 & Q5\_K\_M & 5186 & 7510 & 4.0 & 5.0 \\
 & Q4\_K\_M & 3561 & 10849 & 3.0 & 6.0 \\
 & Q3\_K\_M & 5227.5 & 6693 & 4.0 & 5.0 \\
 & Q2\_K   & 4010 & \textbf{1506.5} & 3.0 & \textbf{1.0} \\
\bottomrule
\end{tabular}
\end{table}

At every precision in both Bieliks except 11B Q2\_K, the median failing trajectory costs more tokens than the median passing one, by factors of 1.18 to 7.65; at 11B Q2\_K the failures are the cheaper ones (medians of 1{,}506.5 against 4{,}010 tokens). The two models part company on how they fail. At 2-bit the 7B fails long and the 11B fails short: the 7B's median failing trajectory is 9{,}139 tokens over 4 steps (interquartile range 6{,}076 to 23{,}427 tokens, 3 to 8 steps) against the 11B's 1{,}506.5 tokens over one step (1{,}126.5 to 5{,}117.5 tokens, 1 to 4 steps); at 3-bit, one precision above the cliff, the 7B's is 26{,}833 tokens over exactly the 8-step budget (11{,}790 to 29{,}742), against 3{,}509 tokens over 3 steps when it succeeds, while the 11B's is 6{,}693 tokens over 5 steps. At 3-bit, 27 of the 7B's 36 failures exhaust the task's step budget and 9 contain a parsed \texttt{final\_answer}; at 2-bit only 19 of 57 exhaust the budget and 38 contain one. The 11B at 2-bit has a parsed \texttt{final\_answer} in 50 of 64 failures and exhausts the budget in 14, and it answers before it has looked: 37 of its 64 failures end at the first step with a well-formed \texttt{final\_answer}, no tool call and no parse error, all but one asserting a temperature, a forecast or a sent alert that the empty tool log never produced. Of those 37, 18 reuse the pattern of the system prompt's example sentence (\emph{W Krakowie jest 7.5°C i pada deszcz.}): 15 retain 7.5°C, including the unchanged \texttt{adv\_005}, and 3 substitute the temperature as well as the city. At the same threshold the child fails longer and mostly still produces a final answer, while the parent usually answers before it looks. Because the pair shares its lineage and tokenizer and differs by pruning, distillation and the child's own post-training, this difference is a property of the compressed checkpoint rather than of architecture, though which of those components carries it is not separable here. What the three-model comparison establishes is negative and useful: \textbf{a failure-mode taxonomy fitted to one model should not be assumed to transfer}, even when the aggregate curve looks similar and the collapse threshold is identical.

\subsection{Envelope collapse: one defect behind the 7B's 8-bit protocol failures}
\label{sec:envelope}
The 7B's dominant 8-bit failure has a single identifiable cause. The agent protocol requires tool invocations of the form \verb|{"action":"call_tool","tool":X,"arguments":{...}}|. On a family of tasks the 7B produces correct calls for the first two tools and then, exactly when a third distinct tool enters the chain, drops the envelope and writes the tool name directly into the \texttt{action} field, as \verb|{"action":"convert_temperature","arguments":{...}}|. The parser rejects this as an unknown action, the tool never executes, and the ordered-sequence oracle records a wrong tool order.

\begin{sloppypar}Across the 7B Q8\_0 main-suite run at \texttt{repair\,=\,off}, counting steps rejected as \texttt{unknown\_action} whose raw output names an available tool in the \texttt{action} field gives 28 steps naming \texttt{convert\_temperature} and 3 naming \texttt{send\_weather\_alert}; a step is counted once per named tool. In the three distinct 3-tool \texttt{EN\_EN} prompts the first collapse occurs at the third step, and their six task slots account for 12 of the conversion steps. Its clearest expression is the set of 3-tool \texttt{EN\_EN} chains, where the 7B at 8-bit scores 0 of 6 task slots, every one an envelope collapse. Two independent lines of evidence show this is a protocol defect rather than an ability limit: the one-shot repair layer, which rewrites the malformed envelope, converts 5 of the 6 slots to passes, and the 11B solves the same slots at 3 to 5 of 6 across healthy precisions (4, 5, 3, 4, 5 from 8-bit down to 3-bit). Three of the six slots are twins of the other three and, in this cell, produce identical trajectories (verified on the released matrix; identity is not guaranteed in general, Section~\ref{sec:mines}), so the effective sample is three prompts, and we run no significance test on it. An adjacent zero cell, the 4-tool \texttt{PL\_EN} chains, has a different signature (missing final answer after 8 steps, no unknown actions), so these are two distinct protocol regressions. This mechanism also explains the repair asymmetry of Section~\ref{sec:discussion}: repair is worth 8 hard-tier tasks to the 7B at 8-bit because its failures there are repairable envelope defects, and worth nothing to the 11B, which does not make that error. Appendix~\ref{app:envelope} gives raw output.\end{sloppypar}

\section{Finding 3 (exploratory): The unscaffolded floor, scaffolding, and the order trap}
\label{sec:f3}

The extended ladder of Section~\ref{sec:ladder} was run on both Bieliks at four precisions (Q8\_0, Q4\_K\_M, Q3\_K\_M, Q2\_K), at both repair settings; PLLuM sits at the floor of the entire ladder ($\le 2/46$ at every precision) and is excluded from ladder inference; for completeness, its only two ladder passes are both at L0 (instances c and f at 8-bit, duplicate inputs and so one distinct prompt), which nominally reverses the gradient direction at $n_{10}=2$, $p=0.5$, far from any inference. Table~\ref{tab:ladder} reports per-rung pass rates at the strict rate, with corrected values where the typing artifact of Section~\ref{sec:mines} changes a cell, and adds the column L0e, the 2026-09 rerun of L0 with the no-tool rule stated in the prompt, available for the four cells at Q8\_0 and Q4\_K\_M (Appendix~\ref{app:ladder}).

\begin{table}[h]
\centering\small
\caption{Extended ladder, pass rate per rung, \texttt{repair\,=\,off}. L0, L0e, L1, L2, L3N have $n=10$; L3T has $n=6$. L0e is the 2026-09 rerun of L0 with an explicit no-tool sentence in the prompt, run at Q8\_0 and Q4\_K\_M only (identical values at \texttt{repair\,=\,on}); a dash marks cells not rerun. Corrected values (format-only failures carrying the gold value forgiven, Section~\ref{sec:mines}) in parentheses where they differ; no other cell changes. L0 is 0.00 in all eight Bielik runs at both rates; L0e is 0.10 in one cell and 0.00 in the other three. The total column sums the five original rungs and excludes L0e; the 11B at Q8\_0 is therefore 13 strict without it.}
\label{tab:ladder}
\setlength{\tabcolsep}{4pt}%
\resizebox{\linewidth}{!}{%
\begin{tabular}{llccccccc}
\toprule
& & L0 (0) & L0e (0, explicit) & L1 (1) & L2 (2) & L3T (4, trap) & L3N (4, no graph) & total /46 \\
\midrule
\multirow{4}{*}{11B}
 & Q8\_0   & 0.00 & 0.10 & 0.10 & 0.40 & \textbf{1.00} & 0.20 (\textbf{0.90}) & 13 (20) \\
 & Q4\_K\_M & 0.00 & 0.00 & 0.00 & 0.10 & 0.00 & 0.00 (0.40) & 1 (5) \\
 & Q3\_K\_M & 0.00 & -- & \textbf{1.00} & \textbf{1.00} & 0.83 & 0.70 & 32 \\
 & Q2\_K   & 0.00 & -- & 0.00 & 0.00 & 0.00 & 0.00 & 0 \\
\midrule
\multirow{4}{*}{7B}
 & Q8\_0   & 0.00 & 0.00 & 0.00 & 0.20 & 0.00 & 0.00 (\textbf{0.70}) & 2 (9) \\
 & Q4\_K\_M & 0.00 & 0.00 & 0.00 & 0.00 & 0.00 & 0.10 (0.30) & 1 (3) \\
 & Q3\_K\_M & 0.00 & -- & 0.10 & 0.10 & 0.00 & 0.00 & 2 \\
 & Q2\_K   & 0.00 & -- & 0.00 & 0.00 & 0.00 & 0.00 & 0 \\
\bottomrule
\end{tabular}%
}
\end{table}

\begin{table}[h]
\centering\small
\caption{Paired exact McNemar on the ladder with the explicit-prohibition rerun L0e as the baseline, instance slot as pairing unit, \texttt{repair\,=\,off}; L0e has ten slots but eight distinct inputs (\texttt{a}=\texttt{e}, \texttt{c}=\texttt{f}). The twelve-row family was fixed at the analysis stage, after the L0e rerun: L0e against L3N at the strict and at the corrected rate, and against L3T, for both Bieliks at Q8\_0 and Q4\_K\_M. $n_{01}$ counts instances failed at L0e and passed at the scaffolded rung, $n_{10}$ the reverse; the last column gives the nominal $p$ and the Holm-adjusted $p$ within the family. Asterisks mark nominal $p<0.05$; no contrast survives correction. The original L0-based contrasts, including those at Q3\_K\_M, and the sensitivity excluding duplicate baseline inputs from both rungs are in Appendix~\ref{app:ladder}.}
\label{tab:laddermcnemar}
\begin{tabular}{llllcccc}
\toprule
model & precision & contrast & rate & $n_{11}$ & $n_{10}$ & $n_{01}$ & $p$ / Holm \\
\midrule
11B & Q8\_0 & L0e vs L3N ($n=10$) & strict    & 0 & 1 & 2 & 1.000 / 1.000 \\
11B & Q8\_0 & L0e vs L3N ($n=10$) & corrected & 1 & 0 & 8 & \textbf{0.0078}* / 0.094 \\
11B & Q8\_0 & L0e vs L3T ($n=6$)  & strict    & 0 & 0 & 6 & \textbf{0.031}* / 0.31 \\
11B & Q4\_K\_M & L0e vs L3N ($n=10$) & strict    & 0 & 0 & 0 & 1.000 / 1.000 \\
11B & Q4\_K\_M & L0e vs L3N ($n=10$) & corrected & 0 & 0 & 4 & 0.125 / 1.000 \\
11B & Q4\_K\_M & L0e vs L3T ($n=6$)  & strict    & 0 & 0 & 0 & 1.000 / 1.000 \\
7B  & Q8\_0 & L0e vs L3N ($n=10$) & strict    & 0 & 0 & 0 & 1.000 / 1.000 \\
7B  & Q8\_0 & L0e vs L3N ($n=10$) & corrected & 0 & 0 & 7 & \textbf{0.016}* / 0.17 \\
7B  & Q8\_0 & L0e vs L3T ($n=6$)  & strict    & 0 & 0 & 0 & 1.000 / 1.000 \\
7B  & Q4\_K\_M & L0e vs L3N ($n=10$) & strict    & 0 & 0 & 1 & 1.000 / 1.000 \\
7B  & Q4\_K\_M & L0e vs L3N ($n=10$) & corrected & 0 & 0 & 3 & 0.250 / 1.000 \\
7B  & Q4\_K\_M & L0e vs L3T ($n=6$)  & strict    & 0 & 0 & 0 & 1.000 / 1.000 \\
\bottomrule
\end{tabular}
\end{table}

Four things are true at once, and stating them separately is the point of the dual-arm design.

\paragraph{The unscaffolded rung is a floor, and the rerun shows what kind.} L0, where the numbers sit in the prompt and the oracle rejects any tool call, is 0 of 10 in all eight Bielik runs at both rates: the typing correction rescues nothing here, because the L0 failures are not mistyped correct answers. The prompt as run, however, never told the model that calls were forbidden; the rule lived only in the oracle (Artifact 3, Section~\ref{sec:mines}), so the original zeros could not distinguish a floor of competence from an unstated instruction. Four of the ten 11B trajectories at Q8\_0 show the difference concretely: instances \texttt{a}, \texttt{b}, \texttt{e} and \texttt{i} (three distinct inputs, since \texttt{a} and \texttt{e} are duplicates) each end in a well-formed final answer carrying the exact gold value ($46.4$, $48.2$, $46.4$ and $42.8$), and each is failed on \texttt{unexpected\_tool\_call} as its only tag, at both repair settings. We therefore reran L0 in 2026-09 with the sentence \emph{Nie używaj żadnych narzędzi. Odpowiedz bezpośrednio.} (``Do not use any tools. Answer directly.'') prepended to every prompt, called L0e, on the four cells at Q8\_0 and Q4\_K\_M for both Bieliks at both repair settings; a control run of the unmodified L0 tasks in the same session reproduced the raw model outputs, token counts and verdicts of the 80 corresponding released L0 trajectories; the control detected no difference in these measures, supporting attribution of the L0e changes to the added sentence (Appendix~\ref{app:ladder}). L0e is 1 of 10 for the 11B at Q8\_0 and 0 of 10 in the other three cells, at both repair settings. The floor stands, and the two models sit on it for different reasons. Under the explicit prohibition the 11B makes no tool call in any of its 20 trajectories, two precisions by ten instances at \texttt{repair\,=\,off} with identical outcomes at \texttt{repair\,=\,on}, and then fails in two different ways, on content at Q8\_0 and on protocol at Q4\_K\_M: at Q8\_0 it averages all four readings instead of the second and fourth in 8 of 10 instances, chiefly an error of selecting the readings. Three answers (\texttt{b}, \texttt{h}, \texttt{i}) also misconvert their stated Celsius value, with \texttt{b} and \texttt{i} among the eight four-reading averages and \texttt{h} using another wrong mean. At Q4\_K\_M it refuses (a JSON \texttt{error} object saying it cannot comply because it does not use tools, scored as \texttt{unknown\_action} or \texttt{no\_json\_found}): six trajectories refuse through the whole eight-step budget, and three refuse for seven steps and state the correct value only at the final step, too late for an oracle that has already logged the protocol errors. The 7B at Q8\_0 disobeys: it calls tools in 10 of 10 trajectories despite the sentence (\texttt{unexpected\_tool\_call} in every failure) and then reaches no schema-valid final answer: all ten trajectories contain a \texttt{final\_answer} with a bare numeric answer. At Q4\_K\_M with repair off it makes no parsed call, but attempts collapsed calls in four trajectories (\texttt{b}, \texttt{h}, \texttt{i}, \texttt{j}); repair on turns those attempts into executed calls. With repair off, four answers average all four readings and five trajectories have no schema-valid final answer. In the original L0 run both models reached for tools at Q8\_0: the 11B used well-formed calls in all ten trajectories, while the 7B made collapsed-envelope attempts in all ten, rejected as \texttt{unknown\_action} with repair off and converted into executed calls with repair on. Under L0e the 11B stops calling, whereas the 7B still calls and now uses well-formed envelopes. One rung, one score, and a third example of Finding~2's lesson that failure modes do not transfer between these two models.

\paragraph{Scaffolding lifts both models off the floor, as an exploratory observation.} With L0e as the baseline, routing the identical computation through four explicit calls takes the 8-bit 11B from 1/10 to 9/10 at the corrected rate ($n_{01}=8$, $n_{10}=0$, exact McNemar $p=0.0078$; excluding duplicate inputs, 1/8 to 7/8, $p=0.031$) and the 8-bit 7B from 0/10 to 7/10 ($n_{01}=7$, $p=0.016$; excluding duplicates, 0/8 to 5/8, $p=0.0625$); at the strict rate the 11B moves only through the trap arm (L3T 6/6 against L0e 0/6, $p=0.031$) and the 7B not at all. The difference between the two rates is the typing artifact for the 11B. For the 7B it is the typing artifact plus a rejected extra call that the correction also forgives: after the four required calls the 7B attempts a fifth conversion that feeds the mean of its two Fahrenheit results back in as Celsius, a unit error written as an unevaluated expression and therefore rejected as \texttt{invalid\_json}, and then returns the right Fahrenheit number as an object instead of a string; because the oracle fails any trajectory with a rejected step, correcting the answer type alone would leave the 7B at 0/10, and its corrected 7/10 is an upper bound that also forgives a unit error the parser happened to block (Appendix~\ref{app:ladder}). The family of twelve contrasts in Table~\ref{tab:laddermcnemar} was fixed at the analysis stage, after the L0e rerun, and after Holm correction none of them survives (adjusted $p=0.094$, $0.17$ and $0.31$ for the three that are nominally significant). Counting all ten slots gives a smallest attainable exact $p$ of $0.002$ with ten discordants and $0.0039$ with nine, against Holm's $0.0042$ threshold for the smallest of twelve. Excluding the duplicate L0e inputs leaves eight paired instances for L3N and four for L3T; even $2/2^8=0.0078$ exceeds that threshold, so no contrast could survive this correction. The ladder is a mechanism probe, not a confirmatory test. We therefore report scaffolding as an exploratory observation: a large, one-directional lift at 8-bit under the corrected reading, which we consider the more informative one because the forgiven 11B trajectories are verifiably correct computations, every forgiven 7B trajectory ends on the gold value, and the control group of substantively wrong answers with identical tags is not forgiven; readers who disagree have the strict numbers. One guard on scope: the defensible statement is that L0e is the floor and that the four-call rungs clear it at 8-bit, where corrected L3N reaches $0.90$ for the 11B and $0.70$ for the 7B and the trap arm is cleared by the 11B alone. It does not hold at 4-bit, where the trap arm is 0 of 6 for both models and corrected L3N reaches only $0.40$ and $0.30$. Nor is it that each added tool helps monotonically, since L3N sits below L2 in several strict cells; the gradient is a contrast between no scaffolding and full scaffolding, not a staircase.

\paragraph{The model split is associated with the order trap.} The two four-call arms share their length and their gold value and differ in whether the chain runs through \texttt{find\_\allowbreak{}nearest\_\allowbreak{}city} and its order trap; they also differ in which tools they call and in where the averaging happens, so the arm contrast isolates the trap only up to those design differences. On the trap-free arm both models succeed at 8-bit at the corrected rate (0.90 against 0.70). On the trap arm they split absolutely at 8-bit: 6 of 6 for the 11B, 0 of 6 for the 7B, at every rate. The pilot ladder's original reading, that the 7B has a flat scaffolding-independent floor, was therefore a compound of two effects the pilot could not distinguish: a genuine inability to survive the order trap, and format-side rejections of an otherwise correct trap-free computation. The extended design separates them. Within the Bielik family, surviving the order trap is a capability the parent has and the pruned-and-distilled child lacks at 8-bit and at 3-bit; at 4-bit and 2-bit both models score 0 of 6 on the trap arm.

\paragraph{The 3-bit 11B beats the 8-bit 11B on light scaffolding.} At Q3\_K\_M the 11B solves L1 and L2 perfectly (10/10 and 10/10) where the 8-bit model manages 1/10 and 4/10, and its 8-bit failures there are substantively wrong values, including the same wrong number returned for five different L1 instances, a degenerate answer the 3-bit model never produces. On the trap-free four-call arm the ordering reverses (corrected 0.90 for Q8 against 0.70 for Q3), and on the main suite Q8 remains ahead overall, so this is not a claim that 3-bit is the better model; it is an illustration that quantization can reorder capabilities non-uniformly across task structure, which no single aggregate number reveals. A single greedy decode underlies every cell here, and Section~\ref{sec:variance} shows that run-level outcomes at Q3\_K\_M vary widely under sampling, so we treat this inversion as a property of the configuration measured, not as a stable ranking.

\subsection{Matched twins: the value penalty, and its robustness to the typing artifact}
\label{sec:twins-results}
The ten matched twin pairs of the main suite give an independent view of value correctness (Table~\ref{tab:twins}). The 7B never exceeds $0.20$ on the value-checked twins at any precision; the 11B reaches $0.80$ to $0.90$ from 8-bit to 5-bit, $0.60$ at the 4-bit dip and $0.70$ at 3-bit. Read at face value, requiring the value costs between $0$ and $0.20$ and helps once, at the 11B's 4-bit point. The decoding divergence of Section~\ref{sec:mines} changes how the table should be read. A twin pair can disagree for two reasons: the trajectories coincide and the value is wrong, which is the penalty the design was built to measure, or the two greedy decodes diverge and one member fails for reasons unrelated to the value. Across the ten cells of Table~\ref{tab:twins}, judging coincidence on the raw model output, the first kind occurs in 5 pairs (4 for the 11B, 1 for the 7B), every one a structure pass with a wrong final value; the second kind occurs in 9 pairs (5 for the 11B, 4 for the 7B; judged on step and token count alone one 7B pair at Q3\_K\_M would move to the first kind). For the 7B all four divergent pairs favour the structure twin, including both discordances at Q4\_K\_M. For the 11B three favour the structure twin (both discordances at Q8\_0 and one at Q4\_K\_M) and two favour the value twin, and those two are exactly the $+0.10$ at Q4\_K\_M: on \texttt{v3\_chain\_004} the structure twin drops the call envelope from the calls it plans ahead in its first-step output and the value twin does not, and \texttt{v3\_chain\_en\_001} fails on invalid JSON in one member only. That entry is a property of the decoder, not of the model at 4-bit, and the defensible statement of the value penalty is the coinciding-trajectory count: small, one-directional, and too sparse for a per-precision test at $n=10$. One robustness note connects this to the typing artifact: a mistyped answer registers no final answer and fails \emph{both} twins, so on coinciding trajectories the artifact cannot manufacture a structure-pass-value-fail discordance. Absolute twin levels share its exposure, and we did not re-audit the main suite for it; that caveat is listed in Section~\ref{sec:limits}.

\begin{table}[h]
\centering
\caption{Matched twin pairs ($n=10$), \texttt{repair\,=\,off}: structure-only pass rate against structure-plus-value. Differences mix a genuine value penalty on coinciding trajectories with disagreements caused by divergent greedy decodes (Section~\ref{sec:mines}); the $+0.10$ at 11B Q4\_K\_M and the $-0.20$ at 11B Q8\_0 are entirely of the second kind. The 7B rows are the four precisions of the original grid; at Q6\_K and Q5\_K\_M the 7B scores $0.200/0.200$ and $0.400/0.200$, both Q5\_K\_M discordances being of the second kind.}
\label{tab:twins}
\begin{tabular}{llccc}
\toprule
model & precision & structure & with value & difference \\
\midrule
\multirow{4}{*}{7B}
 & Q8\_0   & 0.200 & 0.200 & $0.00$ \\
 & Q4\_K\_M & 0.400 & 0.200 & $-0.20$ \\
 & Q3\_K\_M & 0.400 & 0.200 & $-0.20$ \\
 & Q2\_K   & 0.100 & 0.000 & $-0.10$ \\
\midrule
\multirow{6}{*}{11B}
 & Q8\_0   & 1.000 & \textbf{0.800} & $-0.20$ \\
 & Q6\_K   & 1.000 & \textbf{0.900} & $-0.10$ \\
 & Q5\_K\_M & 0.900 & \textbf{0.800} & $-0.10$ \\
 & Q4\_K\_M & 0.500 & 0.600 & $+0.10$ \\
 & Q3\_K\_M & 0.900 & \textbf{0.700} & $-0.20$ \\
 & Q2\_K   & 0.000 & 0.000 & $0.00$ \\
\bottomrule
\end{tabular}
\end{table}

\section{Finding 4: The language gap tracks degradation or task family}
\label{sec:f4}

Table~\ref{tab:lang} gives both Bieliks across all four length by language cells of the deconfounded design.

\begin{table}[h]
\centering
\caption{Pass rate by chain length and interface language ($n=40$ chains), \texttt{repair\,=\,off}. For the 11B, Polish matches or beats English in the long-chain bucket at every healthy precision except the 4-bit dip ($0.31$ against $0.44$); on short chains the single inversion is at Q3\_K\_M ($0.78$ against $0.89$). For the 7B, the long \texttt{PL\_EN} cell is a floor at every precision.}
\label{tab:lang}
\begin{tabular}{llcccccc}
\toprule
model & bucket ($n$) & Q8\_0 & Q6\_K & Q5\_K\_M & Q4\_K\_M & Q3\_K\_M & Q2\_K \\
\midrule
\multirow{4}{*}{7B}
 & short, PL (9)  & 0.78 & 0.89 & 0.89 & 1.00 & 1.00 & 0.11 \\
 & short, EN (9)  & 0.33 & 0.33 & 0.33 & 0.56 & 0.44 & 0.11 \\
 & long, PL (13)  & \textbf{0.08} & \textbf{0.08} & \textbf{0.31} & \textbf{0.15} & \textbf{0.15} & \textbf{0.00} \\
 & long, EN (9)   & 0.56 & 0.56 & 0.78 & 0.33 & 0.22 & 0.00 \\
\midrule
\multirow{4}{*}{11B}
 & short, PL (9)  & 1.00 & 1.00 & 1.00 & 0.89 & 0.78 & 0.00 \\
 & short, EN (9)  & 0.78 & 0.89 & 0.67 & 0.78 & 0.89 & 0.00 \\
 & long, PL (13)  & 1.00 & 0.92 & 0.92 & \textbf{0.31} & 0.77 & 0.08 \\
 & long, EN (9)   & 0.78 & 0.89 & 0.67 & \textbf{0.44} & 0.56 & 0.00 \\
\bottomrule
\end{tabular}
\end{table}

On the 7B one sees an interface gap that reverses with chain length, with Polish ahead on short chains ($0.78$ against $0.33$ at 8-bit) and English ahead on long ones ($0.08$ against $0.56$). On the 11B the same measurement does not reproduce: Polish matches or exceeds English in the long-chain bucket at 8-bit, 6-bit, 5-bit, and 3-bit, and the English advantage appears in exactly one column, Q4\_K\_M, the dip of Section~\ref{sec:f1}. The per-length breakdown in Appendix~\ref{app:lang} is consistent with a task-family association in the 7B: its long \texttt{PL\_EN} cell is pinned between $0.00$ and $0.31$ across precisions, and that cell \emph{contains} the family of 4-tool order-trap chains, 11 of its 13 tasks, the remaining two (\texttt{v3\_chain\_008} and \texttt{v3\_chain\_009}) being five-call chains that route through the same trap. The cell is therefore dominated by, not identical to, the discriminator isolated by the ladder's trap arm in Section~\ref{sec:f3}. The four cells are matched on length band, not on exact length or task family: the short Polish cell holds nine two-call chains, the short English cell three two-call and six three-call chains, and every long Polish chain runs through the neighbour graph while no long English chain does, so the cells compare task families as much as languages. These observations are consistent with an interface gap associated with degradation in the parent and an unsolved structural family in the compressed child. They do not isolate an interface contribution from task family. A benchmark sampling only a degraded configuration, or only one model, could overgeneralize the observed language association.

\section{Under sampling, collapsed runs carry more format failures}
\label{sec:variance}

The mainline of this paper runs at temperature 0 with greedy decoding, which Section~\ref{sec:mines} shows to be reproducible for most but not all tasks. To probe what sampling adds, we reran the 11B main suite at temperature $0.7$ with three seeds at two precisions. The outcome is bimodal (Table~\ref{tab:variance}): two seeds land near the greedy ($T=0$) result and one collapses, at both precisions; Appendix~\ref{app:variance} gives the seed handling and the per-seed forgiveness figures.

\begin{table}[h]
\centering\small
\caption{11B main suite at $T=0.7$, \texttt{repair\,=\,off}, passes of 67. The greedy ($T=0$) reference is 54 (Q8\_0) and 48 (Q3\_K\_M).}
\label{tab:variance}
\begin{tabular}{lccc}
\toprule
precision & seed 1 & seed 2 & seed 3 \\
\midrule
Q8\_0    & \textbf{8}  & 49 & 48 \\
Q3\_K\_M & \textbf{13} & 43 & 51 \\
\bottomrule
\end{tabular}
\end{table}

\begin{sloppypar}Trajectory inspection associates the collapsed runs with format-tagged failures, subject to the bounds below. The runs start normally and the deficit is concentrated in envelope discipline: malformed action wrappers cascade through subsequent steps. Forgiving every failure that carries a format tag (\texttt{unknown\_action}, \texttt{no\_json\_found}, \texttt{invalid\_json}, \texttt{schema\_violation}, \texttt{final\_answer\_shape\_violation} or \texttt{final\_answer\_missing}, the last being a fixation tag in the taxonomy of Section~\ref{sec:taxonomy}; counting it as content instead leaves both ceilings below unchanged and moves failures from format-only to mixed) places seed 1 at $0.776$ (Q8\_0) and $0.881$ (Q3\_K\_M), but that forgiveness also credits 30 and 29 failures that carry a wrong final answer or another content tag alongside the format tag, so these are ceilings rather than recovered pass rates. Forgiving only format-only failures gives $22/67$ and $30/67$; allowing mixed failures but excluding those with a wrong final answer gives $0.687$ and $0.746$. Applied like for like to the other seeds, the generous ceiling puts seed 1 inside the band of seeds 2 and 3 at Q3\_K\_M ($0.821$ and $0.925$) and below it at Q8\_0 ($0.910$ and $0.955$); the stricter version leaves it below the band at both precisions ($0.806$ to $0.881$ and $0.881$ to $0.925$). A competing hypothesis, that a stuck sampler loops, is not supported at Q8\_0, where the seed with the most repeated content is seed 3 (13.7\% of its steps repeat an earlier output, against 6.3\% for the collapsed seed 1), which scores 48 of 67; at Q3\_K\_M the collapsed seed does repeat most (9.8\% against 2.5\% and 3.6\%), so repetition is not excluded there. With two configurations and three seeds this is an observation, not a finding, and we report it for two reasons. It justifies the choice of $T=0$ for measurement, since a pass rate at temperature is evidently not a stable single number for this model. And it connects the collapsed seed to format-tagged failures, also implicated in the 7B's envelope collapse and the typing artifact; these forgiveness bounds do not establish that format degrades before reasoning. One harness detail is worth recording for reproducers: the sampling seed is supplied per step to the completion call rather than once to the model constructor, so each step's sampling stream restarts; this is uniform across seeds and does not by itself explain why one seed collapses.\end{sloppypar}

\section{Discussion}
\label{sec:discussion}

The findings resolve into one sentence: the location of the cliff is a property of quantization, and almost everything else is a property of the model. That is more useful than either half alone. It licenses one genuinely general deployment rule, namely never to run these models below 3 bits for agentic work, while warning against generalizing the accompanying phenomenology, the failure taxonomy, the role of scaffolding, the language gap, from any single model.

For practitioners the guidance is model-specific beyond the shared rule. For the 11B, 6-bit to 8-bit is the safe band, Q4\_K\_M specifically looks worse than both neighbours on this suite (a nominal dip that is significant against neither neighbour after twin exclusion and Holm correction), and 3-bit is surprisingly strong on this suite ($0.716$, and dominant on lightly scaffolded arithmetic) while sitting one step from the cliff. For the 7B, the curve is flat within noise from 8-bit to 3-bit, so choosing a precision matters less than enabling a repair layer, which is worth $+0.12$ at 8-bit because the model's dominant defect is mechanically repairable. For a model that is weak at tool use to begin with, PLLuM here, precision is not the binding constraint and quantization choices are second-order; capability is first-order, until 2-bit removes even that. On arithmetic over tool outputs, the ladder points the opposite way from the intuition that motivated it, though only at the strength of an exploratory observation: neither Bielik clears the unscaffolded rung even when told not to use tools, both reach the right value through four explicit calls at 8-bit, and both then lose most of those results to rejections, the 11B to a bare-float answer and the 7B to a unit-confused, malformed extra call and an object answer; the practical advice is to embed such computations in explicit tool calls, including an explicit conversion step, and to enforce output typing in the harness.

Our original hypothesis H1, that quantization erodes agentic ability before general quality, was not tested as stated, because no matched general-quality curve was measured for these checkpoints; what the data show is narrower: no agentic curve shows monotone early erosion, and two of the three are non-monotone in different ways.

\paragraph{The repair flag is not noise, and its asymmetry is informative.} A one-shot recovery attempt on a malformed step is worth substantially more to the 7B than to the others: on its hard tier it converts 16 passes to 24 at 8-bit and 19 to 25 at 4-bit, lifting the full-suite 8-bit rate from $0.448$ to $0.567$, while for the 11B the same flag changes nothing at 8-bit or 3-bit. Section~\ref{sec:envelope} supplies the reason. For PLLuM repair adds at most $0.045$ anywhere, consistent with failures that are not format-mechanical. Repair, in other words, is a diagnostic: how much it helps tells you what kind of failures a model makes.

\section{Limitations}
\label{sec:limits}
\begin{itemize}
  \item \textbf{The Bielik pair isolates compression, not scale.} The 7B is a pruned and distilled child of the 11B. Differences between them are attributable to that compression as a package (pruning, distillation and the child's own post-training), whose components are not separable from each other here, and none of which is separable from parameter count by this pair alone.
  \item \textbf{The family axis is carried by a weak model.} PLLuM anchors the architecture-and-pretraining axis, but its absolute tool-use level is low, it failed a pre-specified sanity gate (easy tier $0.400$ at 8-bit against a $0.5$ threshold), and its curve was executed as a post-gate extension, which we disclose. Its contribution is therefore limited to within-model contrasts (the cliff) and its failure profile; we cannot rule out that our Polish agentic prompt style disadvantages its instruction tuning, though no harness fault was found in the checks reported in Section~\ref{sec:setup}.
  \item \textbf{The comparison is not like for like in quantization provenance.} The 11B's six quantizations are pre-built from one community publisher; the 7B's Q8, Q6, Q5, and Q4 are the publisher's official files while Q3 and Q2 are requantized from Q8\_0; PLLuM's come from a third source.
  \item \textbf{The typing artifact is documented, not repaired.} The oracle that produced the decisive runs is left unmodified and affected results are reported at strict and corrected rates. The main-suite twins were not re-audited for the artifact as a whole: on coinciding trajectories their discordance pattern is provably immune to it, and none of the divergent discordances is a mistyped correct value, but their absolute levels are not immune.
  \item \textbf{Failure labels are priority-dominated in specific cells} (Section~\ref{sec:taxonomy}); no 2-bit claim rests on them.
  \item \textbf{Sample sizes.} 67 main task slots holding 54 distinct inputs; ladder rungs of 10 and 6 slots, with eight distinct inputs at L0 and L0e; twin contrasts at $n=10$ never individually significant; the sampling probe is two configurations by three seeds.
  \item \textbf{The ladder is not powered for confirmatory testing after multiplicity correction.} Counting ten slots admits a smallest exact $p$ of $0.002$ (ten discordants) or $0.0039$ (nine). After duplicate exclusion the L0e-based contrasts have at most eight instances, whose minimum $p=0.0078$ exceeds Holm's $0.0042$ threshold for the smallest of twelve contrasts. This family was fixed at the analysis stage, after the L0e rerun; the scaffolding effect of Section~\ref{sec:f3} is therefore reported as exploratory. The original L0 prompt did not state the no-tool rule (Artifact 3); the L0e rerun repairs that for four of the fourteen ladder cells, and the L0-based contrasts at Q3\_K\_M and Q2\_K remain as originally run, as do the three pilot L0 tasks of the main suite (Section~\ref{sec:mines}).
  \item \textbf{Greedy decoding is not fully reproducible within a run.} Determinism was verified on one smoke task across model-load cycles; on identical twin prompts the released trajectories coincide in step and token count in 46 of 60 pairs for the 11B, 32 of 60 for the 7B, and 31 of 60 for PLLuM, and are identical in raw output in 45, 28 and 27. Every cell of the grid is therefore one draw from a decoder whose output depends on more than the prompt, consistent with floating-point nondeterminism in CUDA inference, although the one rerun we have reproduced its 80 trajectories exactly; the sampling probe of Section~\ref{sec:variance} shows how wide the spread becomes when randomness is added deliberately.
  \item \textbf{Twin pairs are not independent at $T=0$}, being strongly correlated draws on one prompt; they enter the main-suite tests as task slots, and Appendix~\ref{app:mcnemar} repeats the decisive contrasts with both members removed.
  \item \textbf{Synthetic deterministic environment}; no tool latency or failure variability.
  \item \textbf{Two open anomalies.} The 11B's nominal Q4\_K\_M dip and the single-seed sampling collapse are reported without mechanisms.
  \item \textbf{The envelope-collapse trigger is not isolated}: third-tool position and the specific tool \texttt{convert\_temperature} cannot be separated on this task set.
\end{itemize}

\section{Conclusion}
We measured agentic tool use under GGUF quantization for three Polish-focused models spanning two axes of variation, compression within a family and a change of pretraining family, at six precisions each, on a deterministic benchmark with a Polish prompt and English schema interface. The collapse threshold between 3-bit and 2-bit replicates in all three models, across a fourfold spread in capability, and is on this evidence a property of 2-bit GGUF quantization rather than of any one model. Nothing else replicates. At the cliff the compressed child fails long, at a median of 9.1k tokens over 4 steps, while its parent answers at the first step with a confabulated final answer; the child's run to the full 8-step budget belongs to 3-bit, one precision higher. The foreign-family model fails on content while speaking the protocol fluently. On the arithmetic ladder the unscaffolded rung is a floor for both Bieliks, and a rerun with the no-tool rule stated in the prompt leaves it standing; the two models sit on it for different reasons, the 11B obeying the prohibition and then selecting the wrong readings while the 7B at 8-bit keeps calling tools regardless; scaffolding lifts both off that floor at 8-bit once format-only failures that carry the right value are forgiven (an upper bound for the 7B), an effect we report as exploratory because it does not survive multiplicity correction at this sample size, and the split between the two models on the four-call arms is associated with the order trap. The interface gap is associated with degraded configurations or with an unsolved task family. Under sampling at $T=0.7$, in the single two-precision probe we ran on the 11B, the collapsed seed has many format-tagged failures; the forgiveness ceilings do not establish that format degrades before reasoning. At 2-bit the same model often keeps the protocol well formed while confabulating the content. Along the way three scoring artifacts penalized correct or unproscribed behaviour: rounding-hostile gold values, strict answer typing, and a no-tool rule stated to the oracle but not to the model. A fourth artifact, priority-driven failure labels, shaped the interpretation of failures without changing pass or fail, and we report every affected result in both strict and corrected form or, for the no-tool rule, with a rerun, as a template for how synthetic agentic benchmarks can disclose their own instruments. The natural next steps are a native, non-distilled model at 7B scale from the Bielik lineage, which would separate pruning-and-distillation from scale, and a stronger foreign-family tool-use model, which would carry the family axis at a competitive absolute level.

\section*{Reproducibility}
\begin{sloppypar}
Models: Bielik-11B-v3.0-Instruct (six pre-built quantizations from one community publisher, \path{DevQuasar/speakleash.Bielik-11B-v3.0-Instruct-GGUF}), Bielik-Minitron-7B-v3.0-Instruct (Q8\_0, Q6\_K, Q5\_K\_M, Q4\_K\_M from the official release \path{speakleash/Bielik-Minitron-7B-v3.0-Instruct-GGUF}; Q3\_K\_M and Q2\_K requantized on the evaluation box from the official Q8\_0 file, sha256 \texttt{8154b315c3637b1bcebe7cb76e30c4}\allowbreak\texttt{219ccb4fae3ec5596eed8bdc14b3b39bd2}, with \texttt{llama-quantize -{}-allow-requantize <Q8\_0.gguf> <out.gguf> Q3\_K\_M} and likewise \texttt{Q2\_K}, from a CPU-only build of llama.cpp (\texttt{-DGGML\_CUDA=OFF -DLLAMA\_CURL=OFF}); those two output files were deleted after the run and neither their hashes nor the llama.cpp commit of that build were recorded, so they are the one part of the grid that cannot be rebuilt bit for bit, the nearest documented anchor being the llama.cpp commit vendored in \texttt{llama-cpp-python} 0.3.19, \texttt{c0159f9}), Llama-PLLuM-8B-instruct (community GGUF, \path{mradermacher/Llama-PLLuM-8B-instruct-GGUF}, \texttt{llama-3} chat template). The sha256 of every public GGUF file used ships as \texttt{docs/gguf\_manifest.tsv}. Inference: \texttt{llama-cpp-python} 0.3.19, prebuilt CUDA-12 wheel, single RTX 4090, full offload, $T=0$, seed 42, 512-token completion cap per step, per-task step budget (main suite 4 to 10, 8 on 29 of 67 tasks; 8 on every ladder task); ChatML for the Bieliks. Greedy decoding verified identical in raw output and parsed action on one smoke task across two model-load cycles; within a run, identical prompts coincide in step and token count in 109 of 180 twin pairs across the three models and are identical in raw output in 100 of 180 (Section~\ref{sec:mines}). Every \texttt{summary.json} carries its source commit hash: \texttt{83db813} (7B main), \texttt{a023c3b} (11B main), \texttt{e584b38} (extended ladder, PLLuM curve, 7B grid completion, sampling probe). Commit messages of the development history were normalized at release, and later commits are as written; a full mapping from the development identifiers above to the released commits, together with tree-hash identity checks, ships as \texttt{docs/commit\_map.tsv}; the file trees are byte-identical. Independent oracle re-evaluation reproduces all Bielik main-suite counts (24 of 24 runs). Suites: 67 main tasks and 46 ladder tasks, all validated offline by ideal-trajectory and give-up-trajectory checks before any GPU run; 204 unit tests green at the stamped commits. Rerun of 2026-09 (Section~\ref{sec:f3}, Appendix~\ref{app:ladder}): the L0 rung with the no-tool rule stated in the prompt, plus a same-session control of the unmodified L0 tasks, 16 runs on one RTX 4090 (driver 570.211.01, CUDA 12.8; the original runs used 560.35.03 and 12.6) with \texttt{llama-cpp-python} 0.3.19 from the cu124 wheel \texttt{llama\_cpp\_python-0.3.19-cp312-cp312-linux\_x86\_64.whl} (sha256 \texttt{69d3bd42\ldots97d1702}) and the same four public GGUF files verified by sha256; the control reproduced the raw model outputs, token counts and verdicts of the 80 corresponding original L0 trajectories (latencies differ). Its manifest, trajectories, matrices and analysis ship under \texttt{runs/L0e\_2026-09/}, its task files under \texttt{tasks/ladder\_l0e} and \texttt{tasks/ladder\_l0\_control}. The \texttt{constraints} field of the task files is documentation and is never passed to the model. Statistics: paired McNemar exact (two-sided), instance-paired on the ladder; percentile bootstrap, 10{,}000 resamples, paired over tasks for differences. Released artifacts include task definitions, all trajectories, per-task outcome matrices, the failure classifier, and analysis scripts; the repository is public at \url{https://github.com/JakubPrejzner/polagentbench}. Implementation, experiment execution, and manuscript drafting were assisted by AI systems (Anthropic's Claude; OpenAI's GPT for audits and some draft edits) operating under the author's direction; all design decisions, data judgments, and final content are the author's responsibility.
\end{sloppypar}

\appendix

\FloatBarrier
\Needspace*{0.45\textheight}
\section{Easy-tier per-task outcomes, Bielik pair}
\label{app:easy}

\begin{table}[h]
\centering\small
\caption{Easy tier, \texttt{repair\,=\,off}. P denotes pass, F fail. PLLuM easy-tier totals appear in Appendix~\ref{app:split}.}
\label{tab:easytasks}
\begin{tabular}{llcccc|cccccc}
\toprule
& & \multicolumn{4}{c|}{7B} & \multicolumn{6}{c}{11B} \\
task & probe & Q8 & Q4 & Q3 & Q2 & Q8 & Q6 & Q5 & Q4 & Q3 & Q2 \\
\midrule
adv\_001  & language leakage        & P & P & P & F & P & P & P & P & P & F \\
adv\_002  & language leakage        & P & P & P & P & P & P & P & P & P & F \\
adv\_003  & inflection mismatch     & P & P & P & P & P & P & P & P & P & F \\
adv\_004a & diacritic corruption    & P & P & P & P & P & P & P & P & P & F \\
adv\_004b & final-answer shape      & F & P & P & F & P & P & F & P & F & P \\
adv\_005  & hallucinated result     & P & P & P & P & F & P & F & F & P & F \\
adv\_006  & unauthorized side effect& P & P & P & F & P & P & P & P & P & F \\
adv\_007  & recovery, not found     & P & P & P & P & P & P & P & P & P & P \\
adv\_008  & loop to step budget     & P & P & P & P & P & P & P & P & P & F \\
adv\_009a & arithmetic, two readings& P & P & F & F & P & P & P & P & F & F \\
adv\_009b & wrong discriminator     & P & P & P & F & P & P & P & P & P & F \\
adv\_010  & no tools, final answer  & P & P & F & F & P & P & P & F & P & F \\
adv\_011  & numeral parsing in args & P & P & P & P & P & P & P & P & P & F \\
adv\_012  & conditional negation    & P & P & P & P & P & P & P & P & P & F \\
adv\_013  & ordered 3-tool chain    & P & P & P & F & P & P & P & P & F & F \\
\midrule
\multicolumn{2}{l}{total of 15} & 14 & 15 & 13 & 8 & 14 & 15 & 13 & 13 & 12 & 2 \\
\bottomrule
\end{tabular}
\end{table}

Note: the 7B's Q6\_K and Q5\_K\_M easy-tier totals are 14 and 13 respectively; per-task rows for those precisions are in the released matrix.

\FloatBarrier
\Needspace*{0.45\textheight}
\section{Tier-resolved counts and the repair flag}
\label{app:split}

\begin{table}[h]
\centering\small
\caption{7B tier-resolved pass counts. Repair acts on the hard tier from 8-bit down to 4-bit ($+8$, $+7$, $+5$, $+6$ tasks), does nothing on either tier at 3-bit, and at 2-bit acts on the easy tier only ($+2$).}
\label{tab:split7b}
\begin{tabular}{lccccccc}
\toprule
precision & repair & easy /15 & hard /52 & all /67 & easy rate & hard rate & all rate \\
\midrule
Q8\_0   & off & 14 & 16 & 30 & 0.933 & 0.308 & 0.448 \\
Q8\_0   & on  & 14 & 24 & 38 & 0.933 & 0.462 & 0.567 \\
Q6\_K   & off & 14 & 17 & 31 & 0.933 & 0.327 & 0.463 \\
Q6\_K   & on  & 14 & 24 & 38 & 0.933 & 0.462 & 0.567 \\
Q5\_K\_M & off & 13 & 22 & 35 & 0.867 & 0.423 & 0.522 \\
Q5\_K\_M & on  & 13 & 27 & 40 & 0.867 & 0.519 & 0.597 \\
Q4\_K\_M & off & 15 & 19 & 34 & 1.000 & 0.365 & 0.507 \\
Q4\_K\_M & on  & 15 & 25 & 40 & 1.000 & 0.481 & 0.597 \\
Q3\_K\_M & off & 13 & 18 & 31 & 0.867 & 0.346 & 0.463 \\
Q3\_K\_M & on  & 13 & 18 & 31 & 0.867 & 0.346 & 0.463 \\
Q2\_K   & off & 8  & 2  & 10 & 0.533 & 0.038 & 0.149 \\
Q2\_K   & on  & 10 & 2  & 12 & 0.667 & 0.038 & 0.179 \\
\bottomrule
\end{tabular}
\end{table}

\begin{table}[h]
\centering\small
\caption{11B and PLLuM tier-resolved pass counts. Each cell reads \texttt{repair\,=\,off}\,/\,\texttt{on}, the same convention as Table~\ref{tab:curves}, whose second number each \emph{all /67} row reproduces. Every task repair adds to the 11B lands on the hard tier, at most 4 of them (at 4-bit) and none at all at 8-bit or 3-bit; for PLLuM the opposite holds, 9 of its 11 added tasks being easy-tier.}
\label{tab:split11b}
\begin{tabular}{lcccccc}
\toprule
& Q8\_0 & Q6\_K & Q5\_K\_M & Q4\_K\_M & Q3\_K\_M & Q2\_K \\
\midrule
11B easy /15 & 14\,/\,14 & 15\,/\,15 & 13\,/\,13 & 13\,/\,13 & 12\,/\,12 & 2\,/\,2 \\
11B hard /52 & 40\,/\,40 & 41\,/\,42 & 37\,/\,38 & 23\,/\,27 & 36\,/\,36 & 1\,/\,3 \\
11B all /67  & 54\,/\,54 & 56\,/\,57 & 50\,/\,51 & 36\,/\,40 & 48\,/\,48 & 3\,/\,5 \\
\midrule
PLLuM easy /15 & 6\,/\,7 & 5\,/\,7 & 4\,/\,6 & 8\,/\,10 & 9\,/\,11 & 1\,/\,1 \\
PLLuM hard /52 & 7\,/\,8 & 7\,/\,7 & 3\,/\,3 & 7\,/\,7 & 6\,/\,7 & 0\,/\,0 \\
PLLuM all /67  & 13\,/\,15 & 12\,/\,14 & 7\,/\,9 & 15\,/\,17 & 15\,/\,18 & 1\,/\,1 \\
\bottomrule
\end{tabular}
\end{table}

\FloatBarrier
\Needspace*{0.45\textheight}
\section{Failure taxonomy audit}
\label{app:taxonomy}

\begin{table}[h]
\centering\small
\caption{Tag co-occurrence among failures on the 25 arithmetic-graded tasks, Bielik pair, \texttt{repair\,=\,off}. Every co-occurring case is resolved by rule order rather than by evidence. The three pair columns count \emph{inclusively}: a failure carrying tags from all three classes is counted in all three columns, so the columns overlap and need not sum to the failure total. The same inclusive rule produces the PLLuM counts quoted in Section~\ref{sec:taxonomy}, Section~\ref{sec:f2}, and the paragraph below; counting exclusively, meaning a class set of exactly two, would lower PLLuM's arithmetic-plus-fixation figure from 90 to 76.}
\label{tab:cooc}
\begin{tabular}{llccccc}
\toprule
model & precision & failures & arith+protocol & arith+fixation & protocol+fixation & single category \\
\midrule
\multirow{6}{*}{7B}
 & Q8\_0   & 21 & 3 & 3  & 18 & 1 \\
 & Q6\_K   & 21 & 7 & 5  & 15 & 0 \\
 & Q5\_K\_M & 21 & 3 & 4  & 14 & 4 \\
 & Q4\_K\_M & 21 & 9 & 6  & 13 & 5 \\
 & Q3\_K\_M & 20 & 0 & 2  & 5  & 13 \\
 & Q2\_K   & 25 & 8 & 10 & 12 & 7 \\
\midrule
\multirow{6}{*}{11B}
 & Q8\_0   & 11 & 2 & 3  & 7  & 3 \\
 & Q6\_K   & 10 & 1 & 4  & 4  & 3 \\
 & Q5\_K\_M & 11 & 1 & 3  & 5  & 4 \\
 & Q4\_K\_M & 17 & 2 & 6  & 11 & 2 \\
 & Q3\_K\_M & 10 & 0 & 2  & 0  & 8 \\
 & Q2\_K   & 25 & 4 & 18 & 8  & 3 \\
\bottomrule
\end{tabular}
\end{table}

PLLuM is the least determinate of the three once co-occurrence is counted: 39 of its 140 failures across the curve trigger a single rule branch (per precision 6/23, 9/23, 6/24, 4/23, 9/22, 5/25). The dominant combination, a wrong final answer together with fixation tags, covers 90 of 140 (\texttt{wrong\_tool\_order} 85, \texttt{unexpected\_tool\_call} 5), of which 14 also carry a protocol tag; the as-run priority labels all of these genuine arithmetic, and reversing the priority would relabel 101 of 140. Shape-only tag sets never occur. Three further audit results: two tags nominally in the fixation set are never emitted by the oracle in any clean run, so fixation rests in practice on three tags; the rounding-artifact window was checked against every failure in all runs and matched exactly one, the 7B at Q5\_K\_M on \texttt{v3\_arith\_L3\_a}, whose answer of $46.375$ falls inside the $0.06$ window around the gold $46.4$ although it is itself a miscalculation (the model averages $7.5$ and $8.0$ to $7.75$\,°C, which is $45.95$\,°F), so the hit is coincidental and the rounding hazard proper recurs nowhere (the same trajectory recurs at \texttt{repair\,=\,on}), so the zero reported for every other cell is a measured result and not an untested assumption; and the oracle scores the first parsed final answer while the classifier reads the last, a latent inconsistency that never binds because no failing trajectory emits more than one final answer.

\FloatBarrier
\Needspace*{0.45\textheight}
\section{Paired McNemar contrasts, main suite}
\label{app:mcnemar}

\begin{table}[h]
\centering\small
\caption{7B, \texttt{repair\,=\,off}. The plateau Q8\_0 through Q3\_K\_M shows no significant contrast; every contrast against Q2\_K is significant.}
\label{tab:mcnemar7b}
\begin{tabular}{llccccc}
\toprule
pair & tier & $n_{11}$ & $n_{10}$ & $n_{01}$ & $n_{00}$ & $p$ \\
\midrule
Q8 vs Q4 & all  & 26 & 4 & 8 & 29 & 0.388 \\
Q8 vs Q3 & all  & 23 & 7 & 8 & 29 & 1.000 \\
Q8 vs Q6 & all  & 30 & 0 & 1 & 36 & 1.000 \\
Q5 vs Q4 & all  & 28 & 7 & 6 & 26 & 1.000 \\
Q4 vs Q3 & all  & 27 & 7 & 4 & 29 & 0.549 \\
Q8 vs Q2 & hard & 1  & 15& 1 & 35& \textbf{0.0005} \\
Q8 vs Q2 & all  & 9  & 21& 1 & 36& \textbf{$<$0.0001} \\
Q4 vs Q2 & all  & 10 & 24& 0 & 33& \textbf{$<$0.0001} \\
Q3 vs Q2 & hard & 2  & 16& 0 & 34& \textbf{$<$0.0001} \\
Q3 vs Q2 & all  & 10 & 21& 0 & 36& \textbf{$<$0.0001} \\
\bottomrule
\end{tabular}
\end{table}

\begin{table}[h]
\centering\small
\caption{11B, \texttt{repair\,=\,off}. Q4\_K\_M is nominally significant against both neighbours (uncorrected, twins counted); every contrast against Q2\_K is significant. Fourteen of the fifteen pairwise contrasts form the analysis family; the omitted one, Q6 vs Q3, is $44/12/4/7$, $p=0.077$.}
\label{tab:mcnemar11b}
\begin{tabular}{llccccc}
\toprule
pair & tier & $n_{11}$ & $n_{10}$ & $n_{01}$ & $n_{00}$ & $p$ \\
\midrule
Q8 vs Q6 & all  & 52 & 2 & 4 & 9 & 0.688 \\
Q8 vs Q5 & all  & 49 & 5 & 1 & 12& 0.219 \\
Q8 vs Q4 & all  & 35 & 19& 1 & 12& \textbf{$<$0.0001} \\
Q8 vs Q3 & all  & 42 & 12& 6 & 7 & 0.238 \\
Q8 vs Q2 & all  & 3  & 51& 0 & 13& \textbf{$<$0.0001} \\
Q6 vs Q5 & all  & 49 & 7 & 1 & 10& 0.070 \\
Q6 vs Q4 & all  & 36 & 20& 0 & 11& \textbf{$<$0.0001} \\
Q6 vs Q2 & all  & 2  & 54& 1 & 10& \textbf{$<$0.0001} \\
Q5 vs Q4 & all  & 32 & 18& 4 & 13& \textbf{0.0043} \\
Q5 vs Q3 & all  & 40 & 10& 8 & 9 & 0.815 \\
Q5 vs Q2 & all  & 2  & 48& 1 & 16& \textbf{$<$0.0001} \\
Q4 vs Q3 & all  & 29 & 7 & 19& 12& \textbf{0.029} \\
Q4 vs Q2 & all  & 2  & 34& 1 & 30& \textbf{$<$0.0001} \\
Q3 vs Q2 & all  & 2  & 46& 1 & 18& \textbf{$<$0.0001} \\
\bottomrule
\end{tabular}
\end{table}

\begin{table}[h]
\centering\small
\caption{PLLuM, decisive contrast, all tasks, \texttt{repair\,=\,off}.}
\label{tab:mcnemarpllum}
\begin{tabular}{lccccc}
\toprule
pair & $n_{11}$ & $n_{10}$ & $n_{01}$ & $n_{00}$ & $p$ \\
\midrule
Q3 vs Q2 & 1 & 14 & 0 & 52 & \textbf{0.00012} \\
\bottomrule
\end{tabular}
\end{table}

\paragraph{Sensitivity without twins.} Removing both members of the ten twin pairs (47 tasks, 32 hard) leaves every cliff contrast significant: 7B Q3 vs Q2 on all tasks $9/16/0/22$, $p=3.1\times10^{-5}$, and on the hard tier $1/11/0/20$, $p=0.00098$; 11B Q3 vs Q2 $2/30/1/14$, $p=3.0\times10^{-8}$; PLLuM Q3 vs Q2 $1/13/0/33$, $p=0.00024$; every 7B contrast above the cliff stays non-significant (all ten pairwise contrasts among Q8\_0 to Q3\_K\_M, $p\ge0.34$), and every contrast against Q2\_K stays significant in all three models (fifteen contrasts, largest $p=0.031$, PLLuM Q5 vs Q2). The 11B's Q4\_K\_M contrasts become $23/10/2/12$, $p=0.039$ against Q5 and $20/5/12/10$, $p=0.14$ against Q3. Holm step-down within each table's family with twins counted gives adjusted $p=0.030$ (Q5 vs Q4) and $0.17$ (Q4 vs Q3); with twins removed, $0.27$ and $0.86$. Cells are $n_{11}/n_{10}/n_{01}/n_{00}$.

\FloatBarrier
\Needspace*{0.45\textheight}
\section{Bootstrap intervals}
\label{app:bootstrap}

\begin{table}[h]
\centering\small
\caption{All-task point estimates with 95\% intervals, \texttt{repair\,=\,off}. Every interval here and in Table~\ref{tab:boot-diff} is reported at a fixed resampling seed of 42. Recomputing under a second seed (20260823) leaves 16 of the 18 cells here identical to three decimals and moves a single endpoint by about $0.015$, one task in 67, in the other two: PLLuM Q8\_0 upper $0.299 \to 0.284$ and PLLuM Q5\_K\_M lower $0.045 \to 0.030$. One endpoint moves the same way in Table~\ref{tab:boot-diff}, the 7B Q3-versus-Q2 upper bound $+0.433 \to +0.418$. No interval changes sign or crosses zero.}
\label{tab:boot-point}
\begin{tabular}{llccc}
\toprule
model & precision & passes /67 & rate & 95\% interval \\
\midrule
\multirow{6}{*}{7B}
 & Q8\_0   & 30 & 0.448 & $[0.328,\ 0.567]$ \\
 & Q6\_K   & 31 & 0.463 & $[0.343,\ 0.582]$ \\
 & Q5\_K\_M & 35 & 0.522 & $[0.403,\ 0.642]$ \\
 & Q4\_K\_M & 34 & 0.507 & $[0.388,\ 0.627]$ \\
 & Q3\_K\_M & 31 & 0.463 & $[0.343,\ 0.582]$ \\
 & Q2\_K   & 10 & 0.149 & $[0.075,\ 0.239]$ \\
\midrule
\multirow{6}{*}{11B}
 & Q8\_0   & 54 & 0.806 & $[0.701,\ 0.896]$ \\
 & Q6\_K   & 56 & 0.836 & $[0.746,\ 0.925]$ \\
 & Q5\_K\_M & 50 & 0.746 & $[0.642,\ 0.851]$ \\
 & Q4\_K\_M & 36 & 0.537 & $[0.418,\ 0.657]$ \\
 & Q3\_K\_M & 48 & 0.716 & $[0.597,\ 0.821]$ \\
 & Q2\_K   & 3  & 0.045 & $[0.000,\ 0.104]$ \\
\midrule
\multirow{6}{*}{PLLuM}
 & Q8\_0   & 13 & 0.194 & $[0.104,\ 0.299]$ \\
 & Q6\_K   & 12 & 0.179 & $[0.090,\ 0.269]$ \\
 & Q5\_K\_M & 7  & 0.104 & $[0.045,\ 0.179]$ \\
 & Q4\_K\_M & 15 & 0.224 & $[0.134,\ 0.328]$ \\
 & Q3\_K\_M & 15 & 0.224 & $[0.134,\ 0.328]$ \\
 & Q2\_K   & 1  & 0.015 & $[0.000,\ 0.045]$ \\
\bottomrule
\end{tabular}
\end{table}

\begin{table}[h]
\centering\small
\caption{Selected paired differences on all 67 tasks, \texttt{repair\,=\,off}.}
\label{tab:boot-diff}
\begin{tabular}{llccl}
\toprule
model & contrast & difference & 95\% interval & excludes zero \\
\midrule
7B  & Q4 vs Q3 & $+0.045$ & $[-0.045,\ +0.149]$ & no \\
7B  & Q3 vs Q2 & $+0.313$ & $[+0.209,\ +0.433]$ & \textbf{yes} \\
11B & Q4 vs Q3 & $-0.179$ & $[-0.328,\ -0.030]$ & \textbf{yes} \\
11B & Q3 vs Q2 & $+0.672$ & $[+0.552,\ +0.791]$ & \textbf{yes} \\
PLLuM & Q3 vs Q2 & $+0.209$ & $[+0.119,\ +0.313]$ & \textbf{yes} \\
\bottomrule
\end{tabular}
\end{table}

\paragraph{Sensitivity without twins.} On the 47 non-twin tasks the 11B Q4 vs Q3 paired difference is $-0.149$ with interval $[-0.319, +0.021]$, which includes zero; the other four differences keep their sign and their verdict: 7B Q4 vs Q3 $+0.064$ $[-0.043, +0.170]$, 7B Q3 vs Q2 $+0.340$ $[+0.213, +0.468]$, 11B Q3 vs Q2 $+0.617$ $[+0.468, +0.766]$, PLLuM Q3 vs Q2 $+0.277$ $[+0.149, +0.404]$. Point estimates shift for every model because the twin tasks are chains the 11B mostly solves and the other two mostly fail (for example 11B Q8\_0 $0.766$ $[0.638, 0.872]$, 7B Q8\_0 $0.553$ $[0.404, 0.702]$, PLLuM Q8\_0 $0.255$ $[0.128, 0.383]$); no point-estimate interval changes whether it includes zero.

\FloatBarrier
\Needspace*{0.45\textheight}
\section{Token and step statistics, Bielik pair}
\label{app:tokens}

\begin{table}[h]
\centering\small
\caption{Token and step distributions by outcome, \texttt{repair\,=\,off}. Step budgets are per task, 4 to 10, and are 8 on 29 of the 67 tasks. Token counts, defined as in Table~\ref{tab:tokens}, are comparable only within the Bielik pair (shared tokenizer).}
\label{tab:tokens-full}
\begin{tabular}{llccccccc}
\toprule
model & precision & subset & $n$ & mean tok. & median tok. & p90 tok. & mean steps & median steps \\
\midrule
\multirow{12}{*}{7B}
 & \multirow{2}{*}{Q8\_0}   & pass & 30 & 4540  & 3504  & 8645  & 3.30 & 3.0 \\
 &                          & fail & 37 & 12346 & 12298 & 17219 & 7.27 & 8.0 \\
 & \multirow{2}{*}{Q6\_K}   & pass & 31 & 4280  & 3515  & 8589  & 3.29 & 3.0 \\
 &                          & fail & 36 & 11141 & 12384 & 14820 & 7.03 & 8.0 \\
 & \multirow{2}{*}{Q5\_K\_M} & pass & 35 & 6674  & 3577  & 13752 & 3.74 & 3.0 \\
 &                          & fail & 32 & 15840 & 14065.5 & 29083 & 6.47 & 8.0 \\
 & \multirow{2}{*}{Q4\_K\_M} & pass & 34 & 4857  & 3501  & 11281 & 3.24 & 3.0 \\
 &                          & fail & 33 & 12920 & 10793 & 21006 & 6.30 & 6.0 \\
 & \multirow{2}{*}{Q3\_K\_M} & pass & 31 & 4766  & 3509  & 9844  & 3.23 & 3.0 \\
 &                          & fail & 36 & 22362 & 26833 & 37766 & 7.22 & 8.0 \\
 & \multirow{2}{*}{Q2\_K}   & pass & 10 & 4158  & 3230.5  & 6368  & 2.40 & 2.0 \\
 &                          & fail & 57 & 13795 & \textbf{9139} & 32161 & \textbf{4.70} & \textbf{4.0} \\
\midrule
\multirow{12}{*}{11B}
 & \multirow{2}{*}{Q8\_0}   & pass & 54 & 6309 & 5914 & 10381 & 4.33 & 4.5 \\
 &                          & fail & 13 & 8473 & 8415 & 12806 & 5.62 & 6.0 \\
 & \multirow{2}{*}{Q6\_K}   & pass & 56 & 6173 & 6378.5 & 9675  & 4.30 & 4.5 \\
 &                          & fail & 11 & 7579 & 7510 & 12856 & 5.09 & 5.0 \\
 & \multirow{2}{*}{Q5\_K\_M} & pass & 50 & 6220 & 5186 & 10691 & 4.30 & 4.0 \\
 &                          & fail & 17 & 8742 & 7510 & 15398 & 5.59 & 5.0 \\
 & \multirow{2}{*}{Q4\_K\_M} & pass & 36 & 5432 & 3561 & 10442 & 3.72 & 3.0 \\
 &                          & fail & 31 & 9580 & 10849& 14254 & 5.77 & 6.0 \\
 & \multirow{2}{*}{Q3\_K\_M} & pass & 48 & 5775 & 5227.5 & 8959  & 4.21 & 4.0 \\
 &                          & fail & 19 & 7819 & 6693 & 12739 & 5.11 & 5.0 \\
 & \multirow{2}{*}{Q2\_K}   & pass & 3  & 5629 & 4010 & 10240 & 3.33 & 3.0 \\
 &                          & fail & 64 & 5386 & \textbf{1506.5} & 14650 & \textbf{2.75} & \textbf{1.0} \\
\bottomrule
\end{tabular}
\end{table}

\FloatBarrier
\Needspace*{0.45\textheight}
\section{Length and language, per chain length, 7B}
\label{app:lang}

\begin{table}[h]
\centering\small
\caption{7B hard-tier chains by exact chain length and interface language, \texttt{repair\,=\,off}. The 4-tool \texttt{PL\_EN} family, the order-trap chains, is the model's persistent failure independent of precision.}
\label{tab:lang7b}
\begin{tabular}{llccccc}
\toprule
tools & language & $n$ & Q8\_0 & Q4\_K\_M & Q3\_K\_M & Q2\_K \\
\midrule
2 & PL\_EN & 9  & 7/9  & 9/9  & 9/9  & 1/9 \\
2 & EN\_EN & 3  & 3/3  & 3/3  & 3/3  & 1/3 \\
3 & EN\_EN & 6  & \textbf{0/6}  & 2/6  & 1/6  & 0/6 \\
4 & PL\_EN & 11 & \textbf{0/11} & 1/11 & 2/11 & 0/11 \\
4 & EN\_EN & 1  & 0/1  & 1/1  & 0/1  & 0/1 \\
5 & PL\_EN & 2  & 1/2  & 1/2  & 0/2  & 0/2 \\
5 & EN\_EN & 7  & 5/7  & 2/7  & 1/7  & 0/7 \\
6 & EN\_EN & 1  & 0/1  & 0/1  & 1/1  & 0/1 \\
\bottomrule
\end{tabular}
\end{table}

The two bold cells fail differently: the 3-tool \texttt{EN\_EN} cell by envelope collapse, largely repairable, and the 4-tool \texttt{PL\_EN} cell by exhausting the step budget with no final answer, not repairable.

\FloatBarrier
\section{Worked example: a 4-tool order-trap chain}
\label{app:example}

\paragraph{Prompt, in Polish.} \emph{``Sprawdź pogodę w Katowicach. Następnie znajdź miasta w promieniu osiemdziesięciu kilometrów od Katowic i dla pierwszego miasta z otrzymanej listy pobierz prognozę na cztery dni. Na koniec podaj średnią temperaturę z drugiego i czwartego dnia tej prognozy w stopniach Fahrenheita.''}

\begin{sloppypar}
\paragraph{Gold specification.} Required sequence: \texttt{get\_weather(Katowice)}, \texttt{find\_nearest\_city(Katowice, 80)}, \texttt{get\_forecast(Kraków, 4)}, \texttt{convert\_temperature} Celsius to Fahrenheit. Accepted answers \texttt{["45.5", "45,5"]}; step budget 8; five tools offered, four required, one distractor.

\paragraph{Three traps in one task.} \texttt{find\_nearest\_city("Katowice", 80)} returns \texttt{["Kraków", "Gliwice", "Sosnowiec", "Bielsko-Biała"]} in graph order, at 80, 25, 15, 50\,km: the first city is the farthest, and a model that silently reinterprets ``first'' as ``nearest'' fails. The numerals \emph{osiemdziesięciu} and \emph{cztery} must be parsed to 80 and 4. Only days 2 and 4 enter the mean. The gold is rounding-robust by construction: day 2 is $7.0$\,°C and day 4 is $8.0$\,°C, giving $7.5$\,°C and $45.5$\,°F exactly.
\end{sloppypar}

\paragraph{A real passing trajectory, 11B at Q8\_0, 6 steps.} The model converts both days separately, obtaining $44.6$ and $46.4$, and averages in the final answer: $(44.6+46.4)/2 = 45.5$°F. The oracle checks an ordered subsequence, so this longer route passes; this observed behaviour is also the template for the ladder's graph-free L3N arm.

\paragraph{The same task on the 7B at Q8\_0.} Eight steps, budget exhausted, no final answer, invalid JSON on the terminal step.

\FloatBarrier
\section{Envelope collapse, raw output}
\label{app:envelope}
Raw output for one affected task, 7B at Q8\_0, abbreviated:

\begin{quote}\small
\begin{tabular}{llp{8.2cm}}
step 0 & ok & \verb|{"action":"call_tool","tool":"get_weather",|\linebreak\verb| "arguments":{"city":"Helsinki"}}| \\[2pt]
step 1 & ok & \verb|{"action":"call_tool","tool":"get_forecast",|\linebreak\verb| "arguments":{"city":"Helsinki","days":4}}| \\[2pt]
step 2 & \textbf{unknown action} & \verb|{"action":"convert_temperature",|\linebreak\verb| "arguments":{"value":2.5,"from_unit":"celsius",...}}| \\[2pt]
step 3 & \textbf{unknown action} & \verb|{"action":"convert_temperature",|\linebreak\verb| "arguments":{"value":3.5,...}}| \\[2pt]
step 4 & ok & \verb|{"action":"final_answer","answer":|\linebreak\verb| "... (2.5+3.5)/2 = 3.0°C * 9/5 + 32 = 36°F."}| \\
\end{tabular}
\end{quote}

The first two calls are well formed; at the third distinct tool the envelope disappears. The prose in the final answer picks the right two days but converts their mean wrongly: $3.0\,^{\circ}$C is $37.4\,^{\circ}$F, not 36, so this trajectory carries an arithmetic failure as well as the protocol one (its value-checked twin is tagged \texttt{wrong\_final\_answer} too); the protocol failure is the one that recurs across the family and the one the repair layer removes. On sibling prompts the model additionally emits an arithmetic expression as a bare JSON literal and, after numeric repair, an object where the schema requires a string; the repair layer converts 5 of 6 slots to passes.

\FloatBarrier
\section{Extended ladder detail and the typing artifact}
\label{app:ladder}

\paragraph{Corrected cells.} The typing correction changes exactly four cells, all on L3N: 11B Q8\_0 $0.20 \to 0.90$, 11B Q4\_K\_M $0.00 \to 0.40$, 7B Q8\_0 $0.00 \to 0.70$, 7B Q4\_K\_M $0.10 \to 0.30$. Totals move accordingly (Table~\ref{tab:ladder}). No L0, L1, L2, or L3T cell changes at any precision.

\paragraph{Artifact payloads.} Typical forgiven payloads:
\begin{quote}\small
\verb|{"answer": 46.4}| \hfill (11B, bare float)

\verb|{"answer": {"average_temperature_fahrenheit": 46.4}}| \hfill (7B, object)
\end{quote}
In every forgiven case the four required calls execute in order, no tool-sequence or content tag is raised, and the payload contains the exact gold value. The eleven forgiven 11B trajectories carry only \texttt{final\_answer\_missing} and \texttt{schema\_violation}. Eight of the nine forgiven 7B trajectories (all seven at Q8\_0, \texttt{e} at Q4\_K\_M) also carry \texttt{invalid\_json} from one or two rejected steps before the answer, a fifth \texttt{convert\_temperature} call that passes the mean of the two Fahrenheit results with \texttt{from\_unit} set to Celsius, written as an unevaluated expression such as \verb|(45.5+47.3)/2| (had it parsed, it would have converted 46.4 a second time, to 115.52). The unit error is the model's: the unforgiven \texttt{g} at Q8\_0 makes the same rejected call, later converts its Fahrenheit mean 35.6 as Celsius in a call that parses, and answers 96.08. Since the oracle fails any trajectory with a rejected step, correcting the answer type alone would leave those eight failed; the 7B's corrected rate forgives that step as well and is therefore an upper bound.

\paragraph{Control group, not forgiven.} Fifteen L1 and L2 failures of the 8-bit 11B (nine at L1, six at L2) and sixteen of the 4-bit 11B carry the same \texttt{final\_answer\_missing} and \texttt{schema\_violation} tags but substantively wrong values, for example $59$ where the gold is $46.4$ and $54.2$ where it is $48.2$; all 31 remain failures under the corrected rate; the 7B has three such cases, L2 instances \texttt{a}, \texttt{f} and \texttt{i} at Q8\_0 (mistyped $14.1$, $137.5$ and $13.0$, the same tag set as its forgiven trajectories), which likewise stay failed. The value $59$ recurs across five different L1 instances of the 8-bit 11B (\texttt{a}, \texttt{e}, \texttt{h}, \texttt{i}, \texttt{j}), a degenerate repeated answer.

\paragraph{Shortcut split, empty.} A pre-specified sensitivity split for trajectories that compute correctly but merge the two required conversions into one found zero cases in all eight Bielik ladder runs; the tolerant and strict rates coincide on that dimension.

\paragraph{Statistical notes on the ladder tests.} In the original L0-based data $n_{11}=0$ in all 42 rung comparisons (fourteen cells by L2, L3T and L3N against L0) at both rates: no instance passes both L0 and a scaffolded rung, so each paired table is fixed by its marginals and the exact McNemar reduces to an exact sign test (with L0e as the baseline one instance, \texttt{g} at 11B Q8\_0, passes both L0e and L3N); the pairing is methodologically correct but adds no information beyond the sums. The typing correction is one-directional by construction: none of the 138 L0 failures across the fourteen ladder cells carries only the tags the correction forgives (\texttt{final\_answer\_missing}, \texttt{schema\_violation}, \texttt{invalid\_json}, \texttt{final\_answer\_shape\_violation}), so the correction can only raise scaffolded rungs, never L0, and cannot flip a contrast; we accordingly do not cite the absence of flips as evidence. Finally, $p<0.05$ requires at least six discordant instances in one direction, so the 29 of 42 strict comparisons and 27 of 42 corrected comparisons with zero discordants return $p=1.0$ by definition rather than by measurement.

\paragraph{PLLuM on the ladder.} PLLuM's only two ladder passes across all six precisions are both at L0 (instances c and f at 8-bit, which are duplicate inputs); every scaffolded rung is 0. This nominally reverses the gradient direction at $n_{10}=2$, $p=0.5$, and supports no inference.

\paragraph{L0 mechanisms, original run.} At Q8\_0, Q4\_K\_M and Q3\_K\_M the 11B's ten L0 failures all carry \texttt{unexpected\_tool\_call}: the model reaches for tools although the task supplies the numbers; the prompt did not say that calls were forbidden (Artifact 3). The 7B's ten L0 failures at Q8\_0 carry \texttt{unknown\_action} (10) and \texttt{final\_answer\_missing} (9): with repair off, it attempts collapsed-envelope calls in all ten trajectories, including the invented \texttt{calculate\_average}, but none is parsed as a call. Repair on executes calls in all ten, each then tagged \texttt{unexpected\_tool\_call}. At Q4\_K\_M the 7B also makes collapsed attempts in all ten trajectories, with parsed calls in four at repair off and all ten at repair on. The 11B uses well-formed call envelopes at these three precisions and makes no call in its ten Q2\_K L0 trajectories.

\paragraph{The L0e rerun (2026-09).} The ten L0 tasks were copied with the sentence \emph{Nie używaj żadnych narzędzi. Odpowiedz bezpośrednio.} prepended to the prompt and nothing else changed (\path{tasks/ladder_l0e}; hashes of both sets ship with the repository), and run on the four cells 11B and 7B at Q8\_0 and Q4\_K\_M, \texttt{repair\,=\,off} and \texttt{on}, with the same public GGUF files (sha256 verified), the same runner and oracle, seed 42, $T=0$, and the same \texttt{llama-cpp-python} 0.3.19 cu124 wheel version (the original wheel hash was not recorded, so byte identity cannot be asserted), on an RTX 4090 with driver 570.211.01 and CUDA 12.8 (the original runs: 560.35.03 and 12.6). After each model load the unmodified L0 tasks were run as a control: all 80 control trajectories are identical in raw model output, token count and verdict to the released ones, so the control detected no difference in these measures; latencies differ. Every L0e and control verdict was replayed with the oracle and equals the run log. Results per cell (\texttt{repair\,=\,off}; \texttt{on} has identical pass counts): 11B Q8\_0 1/10 (instance \texttt{g}), no tool call in any trajectory, \texttt{wrong\_final\_answer} in 9, eight of which average all four readings; 11B Q4\_K\_M 0/10, no tool call, \texttt{no\_json\_found} 9 and \texttt{unknown\_action} 9 from refusals citing the prohibition, which run through the whole eight-step budget in six trajectories, \texttt{final\_answer\_missing} 6, and a correct value stated at the final step in three instances (\texttt{a}, \texttt{b}, \texttt{e}); 7B Q8\_0 0/10, tool calls in 10 of 10 trajectories, \texttt{unexpected\_tool\_call} 10, \texttt{schema\_violation} 10, \texttt{final\_answer\_missing} 10; 7B Q4\_K\_M 0/10, no parsed tool call but collapsed attempts in four trajectories, \texttt{final\_answer\_missing} 5, \texttt{wrong\_final\_answer} 5 (four averaging all four readings), \texttt{unknown\_action} 4. At repair on those four trajectories (\texttt{b}, \texttt{h}, \texttt{i}, \texttt{j}) execute calls: \texttt{unexpected\_tool\_call} rises from 0 to 4, \texttt{wrong\_final\_answer} from 5 to 9, and \texttt{final\_answer\_missing} falls from 5 to 1. The original L0-based contrasts are kept in Table~\ref{tab:laddermcnemar-l0} for comparison; the L0e baseline changes both 11B Q8\_0 L3N rows. At the corrected rate $n_{01}$ drops from 9 to 8 because instance \texttt{g} passes both rungs; at the strict rate $n_{10}$ rises from 0 to 1 and $p$ changes from $0.5$ to $1.0$. Post hoc, restricting the family to the four corrected L3N contrasts alone would give Holm-adjusted $p$ of $0.031$ (11B Q8\_0), $0.047$ (7B Q8\_0), $0.25$ and $0.25$; this narrower family is also an analysis-stage choice, and the main text relies on the twelve-contrast family instead.

\begin{table}[h]
\centering\small
\caption{Paired McNemar on the ladder with the original L0 as the baseline, as run in 2026-07, selected contrasts, instance as pairing unit, \texttt{repair\,=\,off}. $n_{01}$ counts instances failed at L0 and passed at the scaffolded rung; $n_{10}=0$ in every row. Holm over these eleven rows leaves 11B Q8\_0 corrected ($0.039$) and 11B Q3\_K\_M L0 vs L2 ($0.021$) below $0.05$.}
\label{tab:laddermcnemar-l0}
\begin{tabular}{llllcc}
\toprule
model & precision & contrast & rate & $n_{01}$ & $p$ \\
\midrule
11B & Q8\_0 & L0 vs L3T ($n=6$)  & strict    & 6 & 0.031 \\
11B & Q8\_0 & L0 vs L3N ($n=10$) & strict    & 2 & 0.500 \\
11B & Q8\_0 & L0 vs L3N ($n=10$) & corrected & 9 & 0.004 \\
11B & Q8\_0 & L0 vs L2 ($n=10$)  & strict    & 4 & 0.125 \\
11B & Q4\_K\_M & L0 vs L3N ($n=10$) & corrected & 4 & 0.125 \\
11B & Q3\_K\_M & L0 vs L2 ($n=10$)  & strict & 10 & 0.002 \\
11B & Q3\_K\_M & L0 vs L3N ($n=10$) & strict & 7 & 0.016 \\
11B & Q3\_K\_M & L0 vs L3T ($n=6$)  & strict & 5 & 0.063 \\
7B  & Q8\_0 & L0 vs L3N ($n=10$) & strict    & 0 & 1.000 \\
7B  & Q8\_0 & L0 vs L3N ($n=10$) & corrected & 7 & 0.016 \\
7B  & Q4\_K\_M & L0 vs L3N ($n=10$) & corrected & 3 & 0.250 \\
\bottomrule
\end{tabular}
\end{table}

\paragraph{Sensitivity excluding duplicate baseline inputs.} L0 and L0e each have eight distinct inputs, with \texttt{a}=\texttt{e} and \texttt{c}=\texttt{f} in prompt, tools and budget. All duplicate L0e pairs have identical raw model outputs at both repair settings; original L0 outputs sometimes diverge despite identical inputs. We remove \texttt{e} and \texttt{f} from both members of every paired contrast, including L3T, leaving $n=8$ for L2 and L3N and $n=4$ for L3T. In the row order of Table~\ref{tab:laddermcnemar}, $n_{01}$ becomes 2, 6, 4, 0, 3, 0, 0, 5, 0, 1, 2, 0; $n_{10}=1$ in the first row and 0 otherwise. The corresponding exact two-sided $p$-values are $1$, $0.03125$, $0.125$, $1$, $0.25$, $1$, $1$, $0.0625$, $1$, $1$, $0.5$, $1$. In the row order of Table~\ref{tab:laddermcnemar-l0}, $n_{01}$ becomes 4, 2, 7, 4, 3, 8, 6, 3, 0, 5, 2 and $n_{10}=0$ throughout; $p$ is $0.125$, $0.5$, $0.015625$, $0.125$, $0.25$, $0.0078125$, $0.03125$, $0.25$, $1$, $0.0625$, $0.5$. No contrast survives Holm correction within its table (smallest adjusted $p=0.375$ and $0.08594$, respectively). With eight distinct inputs the smallest attainable exact $p=2/2^8=0.0078$ already exceeds Holm's $0.0042$ threshold for the smallest of twelve tests, so no contrast could survive the main-text correction even with complete discordance.

\FloatBarrier
\section{Sampling probe detail}
\label{app:variance}
Seed handling: the sampling seed is passed to each step's completion call, not to the model constructor, so every step's sampling stream restarts from the same seed; this is uniform across seeds. Diagnosis of the collapsed seed: trajectories begin normally; the deficit concentrates in malformed action envelopes whose parse failures cascade; forgiving every failure with a format tag (the set defined in Section~\ref{sec:variance}) yields $0.776$ (Q8\_0) and $0.881$ (Q3\_K\_M) for seed 1 (passes 8 and 13, format-only failures 14 and 17, mixed format-and-content failures 30 and 29, all credited), whereas forgiving format-only failures alone yields $22/67$ and $30/67$. Allowing mixed failures but excluding those with \texttt{wrong\_final\_answer} yields $0.687$ and $0.746$. These ceilings associate the collapsed seed with format-tagged failures without establishing that format degrades before reasoning. Under the same forgiveness seeds 2 and 3 reach $0.910$ and $0.955$ at Q8\_0 and $0.821$ and $0.925$ at Q3\_K\_M, so the like-for-like comparison places seed 1 inside their band at Q3\_K\_M and below both at Q8\_0; under the stricter forgiveness seeds 2 and 3 reach $0.881$ and $0.925$ at Q8\_0 and $0.806$ and $0.881$ at Q3\_K\_M, and seed 1 stays below both bands. The repeated-content hypothesis fails at Q8\_0, where seed 3 shows the most repetition (13.7\% of steps) and scores 48 of 67; at Q3\_K\_M the collapsed seed 1 repeats most (9.8\% against 2.5\% and 3.6\%).

\FloatBarrier

\end{document}